\documentclass[lettersize,journal]{IEEEtran}
\usepackage{tikz,xcolor,hyperref}
\definecolor{lime}{HTML}{A6CE39}
\DeclareRobustCommand{\orcidicon}{%
    \begin{tikzpicture}
    \draw[lime, fill=lime] (0,0) 
    circle [radius=0.16] 
    node[white] {{\fontfamily{qag}\selectfont \tiny ID}};    \draw[white, fill=white] (-0.0625,0.095) 
    circle [radius=0.007];    \end{tikzpicture}
    \hspace{-2mm}}
\foreach \x in {A, ..., Z}{%
    \expandafter\xdef\csname orcid\x\endcsname{\noexpand\href{https://orcid.org/\csname orcidauthor\x\endcsname}{\noexpand\orcidicon}}
    }

\usepackage{amsmath,amsfonts}
\usepackage[lined,boxed,ruled, commentsnumbered]{algorithm2e}
\usepackage{array}
\usepackage[caption=false,font=normalsize,labelfont=sf,textfont=sf]{subfig}
\usepackage{textcomp}
\usepackage{stfloats}
\usepackage{url}
\usepackage{verbatim}
\usepackage{graphicx}
\usepackage{cite}
\usepackage{amsmath}
\usepackage{amssymb}
\usepackage{mathtools}
\usepackage{amsthm}
\usepackage{hyperref}
\usepackage{url}
\usepackage[utf8]{inputenc} 
\usepackage[T1]{fontenc}    
\usepackage{hyperref}       
\usepackage{url}            
\usepackage{booktabs}       
\usepackage{amsfonts}       
\usepackage{nicefrac}       
\usepackage{microtype}      
\usepackage{xcolor}         
\usepackage{graphicx}
\usepackage{wrapfig}
\usepackage{multirow}
\usepackage{soul}
\usepackage{makecell}
\usepackage{subfig}
\usepackage{comment}
\usepackage{threeparttable}
\usepackage{float}
\usepackage{bm}
\usepackage[english]{babel}
\def\BibTeX{{\rm B\kern-.05em{\sc i\kern-.025em b}\kern-.08em
    T\kern-.1667em\lower.7ex\hbox{E}\kern-.125emX}}
\usepackage{balance}
\begin{document}
\title{DREAM: Domain-agnostic Reverse Engineering Attributes of Black-box Model}

\author{Rongqing~Li\orcidA{}, Jiaqi~Yu\orcidB{}, Changsheng~Li\orcidC{},~\IEEEmembership{Member,~IEEE,} Wenhan~Luo\orcidD{},~\IEEEmembership{Senior Member,~IEEE,}\\ Ye~Yuan\orcidE{},~\IEEEmembership{Member,~IEEE,}~and~Guoren~Wang\orcidF{}

\thanks{Rongqing Li, Changsheng Li, Ye Yuan, and Guoren Wang are with the School of Computer Science and Technology, Beijing Institute of Technology, Beijing 100081, China (e-mail: lirongqing99@gmail.com; lcs@bit.edu.cn; yuan-ye@bit.edu.cn; wanggrbit@126.com).}

\thanks{Jiaqi Yu is with the Kuaishou Technology, Beijing, China (e-mail: yujiaqi03@kuaishou.com).}

\thanks{Wenhan Luo is with the Hong Kong University of Science and Technology, Clear Water Bay, Hong Kong (e-mail: whluo@ust.hk).}

\thanks{Changsheng Li is the corresponding author.}
}
\IEEEpubid{0000--0000/00\$00.00~\copyright~2021 IEEE}


\maketitle

\begin{abstract}
Deep learning models are usually black boxes when deployed on machine learning platforms. Prior works have shown that the attributes (e.g., the number of convolutional layers) of a target black-box model can be exposed through a sequence of queries. There is a crucial limitation: these works assume the training dataset of the target model is known beforehand and leverage this dataset for model attribute attack. However, it is difficult to access the training dataset of the target black-box model in reality. Therefore, whether the attributes of a target black-box model could be still revealed in this case is doubtful. In this paper, we investigate a new problem of black-box reverse engineering, without requiring the availability of the target model's training dataset. We put forward a general and principled framework DREAM, by casting this problem as out-of-distribution (OOD) generalization. In this way, we can learn a domain-agnostic meta-model to infer the attributes of the target black-box model with unknown training data. This makes our method one of the kinds that can gracefully apply to an arbitrary domain for model attribute reverse engineering with strong generalization ability. Extensive experimental results demonstrate the superiority of our proposed method over the baselines.

\end{abstract}


\section{Introduction}
\label{sec:intro}

\IEEEPARstart{I}{n} \textcolor{black} {recent years, machine learning technology has been widely used in in many tasks such as NLP and areas such as image classification \cite{iscen2023improving,ding2024unireplknet,zhou2023implicit,li2021deep}, natural language processing \cite{lin-etal-2023-linear, tanwar2023multilingual,zhu2023investigating,yang2024give}, and speech recognition \cite{10447952, 10448048}. However, existing machine learning frameworks are complicated, which requires substantial computational resources and efforts for users to train and deploy, especially for the non-expert ones. As a result, with the benefits of usability, and cost efficiency, Machine Learning as a Service (MLaaS) has become popular. MLaaS deploys well-trained machine learning models on cloud platforms, allowing users to interact with these models via the provided APIs, making advanced ML capabilities both accessible and affordable.
}

Generally speaking, the machine learning service deployed on the cloud platform is a black box, where users can only obtain outputs by submitting inputs to the model. 
The model's attributes such as architecture, training set, and training method, are concealed by the provider.
However, a question remains: is the deployment safe? \textcolor{black}{Once the attributes of a model are revealed by an adversary, a black-box model becomes a white-box model, introducing significant security threats. On one hand, white-box models are more vulnerable to various types of attacks compared to black-box models such as adversary example attacks \cite{li2022decision,li2021adversarial,chang2022adversarial,liu2023more,wu2022unified} and model extraction attacks \cite{yan2023explanation,beetham2022dual,carlinistealing,zhao2024fully}. Specifically, adversaries train adversarial examples on a surrogate model with the intention of transferring these examples to the target model. 
Research shows that the transferability of adversarial samples increases if the the surrogate architecture is similar to the target model \cite{liu2017delving}.
In this context, model reverse engineering becomes a powerful tool for adversaries to select a surrogate model.
In addition, model extraction attacks aims to extract the functionality of a target black-box model in a surrogate model. Research shows that better extraction performance is achieved when the surrogate model closely resembles the target black-box model \cite{sha2023can}.
Therefore, model reverse engineering can provide significant insights for selecting the architecture of the surrogate model for the extraction.
On the other hand, intellectual property is jeopardized. After adversaries reveal the attributes of the model, they may replicate a model with similar capabilities for commercial purposes, potentially causing indirect economic losses.}

\begin{figure*}
\setlength{\belowcaptionskip}{-0.4cm} 
  \begin{center}
  \includegraphics[width=1\linewidth]{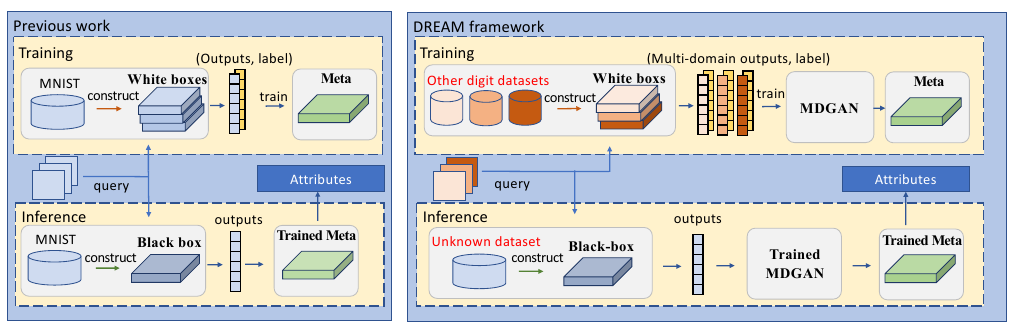}
  \caption{
      Previous work (left) assumes the dataset used to train the target black-box model is known beforehand, and requires to use the same dataset to train white-box models.
      Our DREAM framework (right) relaxes the condition that training data of the target black-box model is no longer required to be available. Our idea is to cast the task of the black-box model attribute inference into an OOD learning problem.
     } 
  \label{fig:intro_fig}
  \end{center}
\end{figure*}

\IEEEpubidadjcol
The work in \cite{oh2018towards2} delves into model reverse engineering to reveal model attributes, as depicted in the left of Fig. \ref{fig:intro_fig}. They first construct a large set of white-box models which are trained based on the same datasets as the target black-box model, $e.g.$, the MNIST hand-written dataset \cite{lecun_mnist_1998}.
Then, the outputs of white-box models are obtained through a sequence of input queries. Finally, a meta-model is trained to learn a mapping between model outputs and model attributes. For inference, outputs of the target black-box model are fed into the meta-model to predict model attributes. The promising results demonstrate the feasibility of model reverse engineering.

However, a crucial limitation in \cite{oh2018towards2} is that they assume the dataset used for training the target model to be known in advance, and leverage this dataset for meta-model learning.

In most application cases, the training data of a target black-box model is unknown. 
When the distribution of training data of the target black-box model is inconsistent with that of the set of constructed white-box models, the meta-model is usually unable to generalize well on the target black-box model.

To verify this point, we train three black-box models with the same architecture on three different datasets, Photo, Cartoon and Sketch\cite{li2017deeper}, respectively. Subsequently, we employ the approach outlined in \cite{oh2018towards2} to train a meta-model using the white-box models trained specifically on the Cartoon dataset
Then, we utilize the trained meta-model to infer attributes of the three black-box models.

\begin{figure}
\setlength{\belowcaptionskip}{-0.4cm} 
  \begin{center}
  \includegraphics[width=0.9\linewidth]{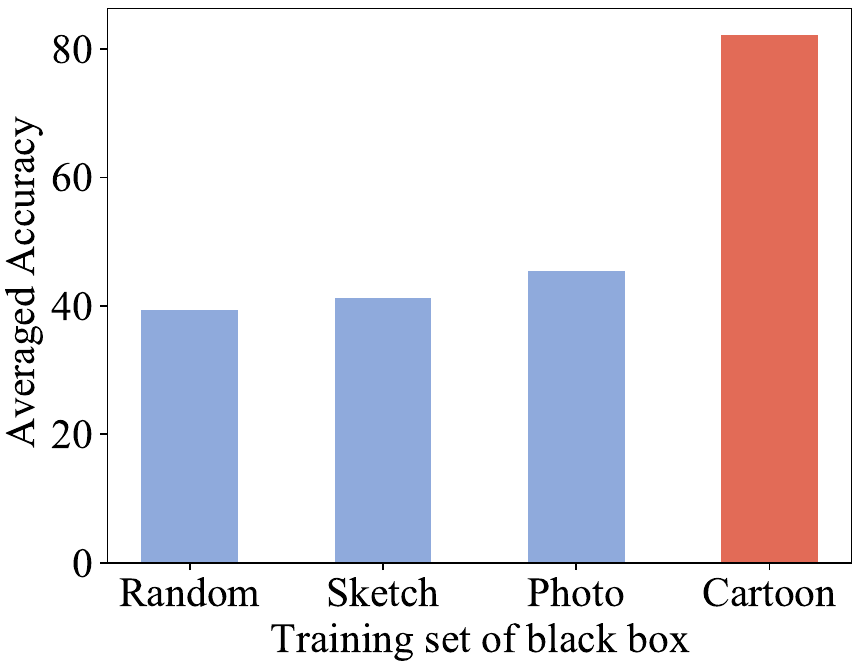}
  \caption{
    The performance of KENNEN\cite{oh2018towards2}
 on black-box model trained on Cartoon, Sketch and Photo dataset \cite{li2017deeper}. The training set of white-box models is Cartoon.
}
  \label{fig:intro_2}
  \end{center}
  \vspace{-2em}
\end{figure}

The results are shown in Fig. \ref{fig:intro_2}. \textcolor{black}{The y-axis represents the average accuracy of reverse engineering for each model attribute (e.g., activation function, optimizer, etc.).}
When the training dataset of black-box models and white-box models are the same (i.e., Cartoon), the performance reaches about $80\%$. Otherwise, it sharply drops to about $40\%$ — close to random guess. This substantial gap underscores the non-trivial nature of investigate model reverse engineering, when the training dataset for the target black-box model is unavailable.


\textcolor{black}{
In this paper, we investigate the problem of reverse engineering the attributes of black-box models without requiring access to the target model's training data.
When the same input queries are fed to models with the same architecture but trained on different datasets, the output distributions typically differ.
Therefore, a key point to our problem setting is bridging the gap between the output distributions of white-box and target black-box models, given the absence of the target model's training data.  An ideal meta-classifier should be well trained based on the outputs of white-box models, and predict well on outputs of the target black-box model, even if white-box and black-box models are trained on different types of data.}

\textcolor{black}{
In light of this, we cast such a problem as an OOD generalization problem, and propose a novel framework DREAM: Domain-agnostic Reverse Engineering the Attributes of black-box Model, as shown in the right of Fig. \ref{fig:intro_fig}. OOD generalization learning has been widely studied in recent years in the field of computer vision, and shows powerful performance \cite{shen2021towards,zhou2022domain,wang2022generalizing}. Its goal is to learn a model on data from one or multiple domains and to generalize well on data from another domain that has not been seen during training. 
One kind of mainstream OOD learning approach is to extract domain-invariant features from data of multiple different domains, and utilize the domain-invariant features for downstream tasks \cite{li_domain_2018,kim2021selfreg,zhou2021mixstyle,li2021learning}. 
Returning to our problem, black-box models deployed on cloud platforms usually provide the label categories of their outputs. Therefore, we can collect data of different styles according to the model's label space as OOD dataset. This overlap ensures that the outputs of the white-box models include similar information with those of the black-box model to some extent. Numerous white-box models are then trained on this OOD dataset, and we obtain outputs by querying these models. These outputs are used as source domains to achieve OOD generalization for reverse engineering black-box models.
}

Since the data we concentrate on is related to the outputs of models, $e.g.$, probability values, how to design an effective OOD generalization method over this type of data has not been explored. 
To this end, we introduce a multi-discriminator generative adversarial network (MDGAN) to learn domain-invariant features from the outputs of white-box models trained on multi-domain outputs.
Based on these learned domain-invariant features, we learn a domain-agnostic reverse meta-model\footnote{\textcolor{black}{The meta-model is used to infer the attributes of a model, thereby "reverse-engineering" a black-box model into a white-box model. Hence, we name it the "reverse meta-model".}}, which is capable of accurately inferring the attributes of the target black-box model trained on unknown data.
 
Our contributions are summarized as follows:
1) We provide the first study on the problem of domain-agnostic reverse engineering the attributes of black-box models and cast it as an OOD generalization problem;
2) We propose a generalized framework, DREAM, to address the problem of inferring the attributes of a black-box model with an unknown training dataset;
3) We constitute the first attempt to explore learning domain-invariant features from probability outputs, as opposed to traditional image;
4) We conduct extensive experiments to demonstrate the effectiveness of our method.

\section{Related Works}
\label{related_works}
\textbf{Reverse Engineering of Model Attributes.} Its goal is to reveal attribute values of a target model, such as model structure, optimization method, hyperparameters, $etc$. Current research efforts focus on two aspects, hardware \cite{yan_cache_2020,hua_reverse_2018,zhu_hermes_2021} and software \cite{oh2018towards2,wang_stealing_2019,rolnick2020reverse,asnani2021reverse}. The hardware-based methods utilize information leaks from side-channel \cite{hua_reverse_2018,yan_cache_2020} or unencrypted PCIe buses \cite{zhu_hermes_2021} to invert the structure of deep neural networks. Software-based methods reveal model attributes by machine learning. \cite{wang_stealing_2019} steals the trade-off weight of the loss function and the regularization term. They derive over-determined linear equations and solve the hyperparameters by the least-square method. 
\textcolor{black}{\cite{asnani2021reverse} theoretically proves the weight and bias can be reversed in linear network with ReLU activation. \cite{rahman_correlation-aware_2020} infers hyperparamters and loss functions of generative models through the generated images.}
KENNEN \cite{oh2018towards2} prepares a set of white-box models and then trains a meta-model to build a mapping between model outputs and their attributes. It is the most related work to ours. However, a significant difference is that KENNEN \cite{oh2018towards2} requires the data used to train the target black-box model to be given beforehand. Our method relaxes this condition, $i.e.$, we no longer require the training data of the target model to be available, which is a more practical problem.

\textbf{Model Functionality Extraction.} It aims to train a clone model that has similar model functionality to that of the target model. To achieve this goal, many works have been proposed in recent years  \cite{orekondy_knockoff_2019,papernot_practical_2017,kariyappa_maze_2021,wang2022dst, yan2023explanation,carlinistealing,zhao2024fully}. \cite{orekondy_knockoff_2019} uses an alternative dataset collected from the Internet to query the target model. \cite{papernot_practical_2017} assumes part of the dataset is known and then presents a dataset augmentation method to construct the dataset for querying the target model. Moreover, data-free extraction methods \cite{kariyappa_maze_2021,papernot_practical_2017,wang2022dst,wang2021delving} query a target model through data generated by a generator, without any knowledge about the training data distribution. Different from the methods mentioned above, our goal is to infer the attributes of a black-box model, rather than stealing the model function.

\textbf{Membership Inference.} Its goal is to determine whether a sample belongs to the training set of a model \cite{he2020segmentations,rezaei2021difficulty,9833649,10.1145/3548606.3560675,10.1145/3548606.3560684,bertran2024scalable,dubinski2024towards,matsumoto2023membership}. Although inferring model attribute is different from the task of membership inference, the technique in \cite{oh2018towards2} is similar to those of membership inference attack. However, as stated aforementioned, when the domain of training data of the target black-box model is inconsistent with that of the set of white-box models, the method is usually unable to generalize well because of the OOD problem.

\textbf{Out-of-distribution Generalization.} 
\textcolor{black}{ Machine learning models often suffer from performance degradation during testing when the distribution of the training data (i.e., the source domains) differs from the test data distribution (i.e., the target domain). This is an out-of-distribution (OOD) problem. \cite{shen2021towards}. One straightforward approach is to leverage data from target domain to adapt the model trained on the source domains. This method is known as OOD adaptation, which has been successfully applied in image and video related tasks \cite{mattolin2023confmix,wang2023ssda3d,mekhazni2023camera,wei2024unsupervised}. However, in many application scenarios, data from the target domain is difficult to obtain. Therefore, OOD generalization methods are introduced \cite{dayal2024madg,lin2024diversifying,zhang2024domain,wang2023sharpness,zhou2024mixstyle}, aiming to train a model by utilizing data from several source domains so that it can generalize well to any unseen OOD target domain.
Existing OOD generalization methods mainly fall into three categories: invariant learning \cite{kim2021selfreg,li_domain_2018,zhou2021mixstyle,zhou_domain_2021,hu_domain_2020}, causal learning \cite{arjovsky2019invariant,creager2021environment,krueger2021out,mahajan2021domain}, and stable learning \cite{shen2020stable,kuang2020stable,zhang2021deep,kuang2018stable}. Invariant learning seeks to minimize the differences among source domains to learn domain-invariant representations. Causal and stable learning aim to identify causal features linked to ground-truth labels from the data and filter out features unrelated to the labels. The former ensures the invariance of existing causal features, while the latter emphasizes effective features strongly related to labels by reweighting attention. Since the above methods primarily focus on images or videos, the design of an effective OOD learning method for attribute inference of black-box models has not been explored so far.}


\section{Proposed Methods}
\begin{figure*}
  \begin{center}
  \includegraphics[width=1.0\linewidth]{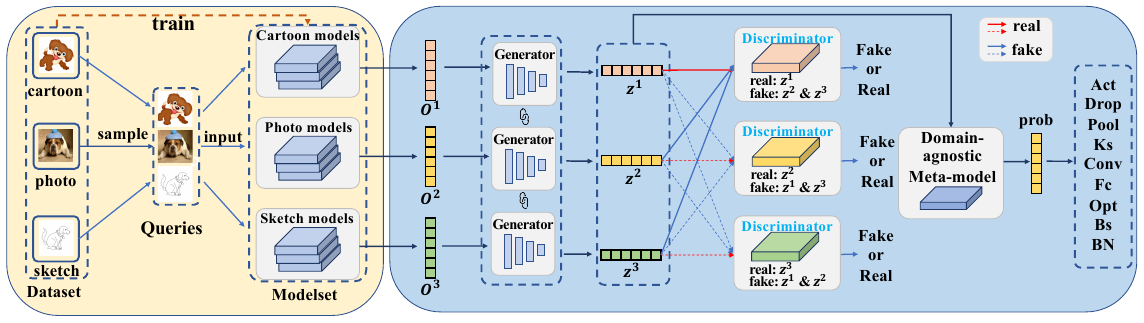}
  \caption{ An illustration of our DREAM framework.
      In the left part, we train a large number of white-box models using datasets collected from different styles (cartoon, photo, and sketch) to construct modelset. Models in the modelset consist of numerous combinations of attributes. Then, we sample queries from each style of dataset and input them into each white-box model to obtain the multi-domain model's outputs $O$.
      In the right part, we propose a multi-discriminator GAN to learn domain-invariant features from the outputs of the white-box models. After that, the domain-agnostic reverse meta-model is trained based on these domain-invariant features. During the inference stage, queries are sent to the black-box model to obtain its outputs. Then, the Generator produces domain-invariant features, which are input to the domain-agnostic meta-model to infer the attributes of the black-box model.
  }
  \label{overall}
  \end{center}
\end{figure*}
\textcolor{black}{
In this section, we first introduce techinical background, threat model and problem formulation in Sect. \ref{sec:preliminaries}. Next, we describe the overall framework of our proposed DREAM in Sect. \ref{sec:overall}. Subsequently, we delve into each components of DREAM in Sect. \ref{sec:obtain} to \ref{sec:metamodel}. Finally, we introduce the training procedure in Sect. \ref{sec:training}.
}

\subsection{Preliminaries}
\label{sec:preliminaries}
\textcolor{black}{We first introduce the background of the KENNEN method. Based on KENNEN, we present the threat model of the problem addressed in this paper. Finally, we introduce the problem formulation.}

\textbf{Background of KENNEN \cite{oh2018towards2}.}
\textcolor{black}{Given a black-box model $B$, model attribute reverse engineering in \cite{oh2018towards2} aims to build a meta-model $\Phi: O \rightarrow Y$, where $O=B(Q)$ is the outputs by querying the black-box model with queries $Q$, and $A$ is the set of model attributes including model architecture, optimizer, and training hyperparameters, $etc$.} 
Concretely, they first construct a large set of white-box models $\mathcal{F}$ containing different attributes combinations and train these white-box models based on the same training data $\mathcal{D}$ as that of the target black-box model. Then outputs $O$ are obtained by querying these white-box models with a sequence of input image queries $Q$. Finally, they train a meta-model $\Phi$ to build mappings from outputs $O$ to model attributes $Y=\{y_k^v|k=1...K, v=1...N^k\}$, where the subscript $k$ represents the type of attributes ($e.g.$, Activation, Dropout), while the superscript $v$ represents the value of the
 attributes ($e.g.$, ReLU/Tanh for Activation, Yes/No for Dropout).
At the inference phase, the meta-model takes outputs from the target model as input and predicts the corresponding attributes.

\textcolor{black}{
\textbf{Threat Model.}
Following KENNEN, we assume that attackers are permitted to query the model and can only access the probability outputs of the model, while attributes such as model structures and optimizers are unable to access. However, KENNEN makes a strong assumption that the training dataset of the model is known. In most cases, obtaining the training dataset is typically challenging. Therefore, we relax this assumption and consider a scenario where only the label space of the black-box model are known. This is a reasonable assumption because a black-box model deployed on the cloud platform typically provides information about its functionality and the categories it can output. Consequently, we can collect data with overlapping label spaces with the black-box target model. This overlap ensures that the outputs of the white-box models include similar information with those of the black-box model to some extent, thereby assisting in learning informative invariant features for achieving attributes reverse engineering.
}

\textbf{Problem Formulation.} As aforementioned, there is a strict constraint in  \cite{oh2018towards2} that they assume the training dataset $\mathcal{D}$ of the target model to be given in advance, and leverage model outputs $O$, where the models are trained on $\mathcal{D}$ for the learning of meta-model $\Phi$. It is difficult to access the training data of a target black-box model, which significantly limits the applications of \cite{oh2018towards2}.
\textcolor{black}{
To mitigate this problem, we provide a new problem setting by relaxing the above constraint, $i.e.$, we no longer require the training data $\mathcal{D}$ of the target black-box model to be available, but only the label space of the black-box model. Consequently, we cast this problem as an OOD generalization problem. To address this, we first collect data $\{D^i\}_{i=1}^M$ from $M$ different sources to train the white-box models. Although these data may contain different styles, all of them include an overlapping label space with the target black-box model. Then, the outputs obtained from the white-box models are divided into $M$ source domains according to the source of the training data for each model. In addition, the outputs from black-box model are the target domain.
Our goal is to leverage outputs from source domains to train a domain-agnostic meta model $\Phi$ that can well generalize to the outputs target domain, thereby enabling it to predict attributes for the target black-box model $B$.
}


\subsection{DREAM Framework}
\label{sec:overall}
To perform domain-agnostic black-box model attribute reverse engineering, we cast this problem into an OOD generalization learning problem, and propose a novel framework DREAM, as shown in Fig. \ref{overall}.
Our DREAM framework consists of two parts:

In the left part of Fig. \ref{overall}, we employ datasets from different domains to train numerous white-box models with diverse attributes, thereby constructing modelsets. (please refer to Sect. \ref{subsec:dataset_construction} for more details). 
Next, we sample queries as input to these models.
For each domain, we sample an equal number of images from the corresponding dataset and concatenate them as a batch of queries. These queries are sent to models belonging to the modelsets. The resulting multi-domain outputs $\{O^i\}_{i=1}^M$ are fed into the subsequent module of our DREAM framework. To learn domain-invariant features, we introduce a MDGAN, as shown in the right part of Fig. \ref{overall}. MDGAN consists of multiple discriminators corresponding to different domains and one generator across multiple domains. The generator is designed to embed model outputs from different domains as features, while each discriminator strives to align the learned feature distributions from other domains with the feature distribution of the domain it corresponds to. In this way, the generator is capable of learning domain-invariant features. Based on the learned domain-invariant features, we further learn a domain-agnostic meta-model to infer the attributes of a black-box model with an unknown domain.

\subsection{Multi-domain Outputs Obtaining}
\label{sec:obtain}
\textcolor{black}{
To achieve model reverse engineering, we first need to obtain the multi-domain outputs of the white-box models. These outputs serve as features of  white-box models' attributes and will be used as inputs for the subsequent modules.} We sample an equal number of images from the source dataset, resulting in queries $Q = \{q_j\}_{j=1}^N$, where $N$ is the number of total queries. 
Subsequently, these queries are fed into the white-box models $\mathcal{F} = [\mathbf{f^1}, \mathbf{f^2}, ..., \mathbf{f^M}]$, where $\mathbf{f^i}$ represents models from the $i^{th}$ domain. 
For each domain $i$, the models $\mathbf{f^i}$ produces outputs $\{O_j^i\}_{j=1}^N \in \mathbb{R}^{N\times C}$, where $C$ is the number of classes in the dataset. We then concatenate the outputs as an 1-dimensional vector $O^i \in \mathbb{R}^{NC}$.
Finally, the multi-domain outputs are derived as $O = \{O^i\}_{i=1}^M$.



\subsection{Multi-Discriminator GAN}
After preparing multi-domain outputs, we introduce MDGAN on the basis of \cite{goodfellow2020generative}. The objective is to learn domain-invariant features from these probability outputs of white-box models trained on different domains.

 \begin{figure}
  \begin{center}
  \includegraphics[width=1\linewidth]{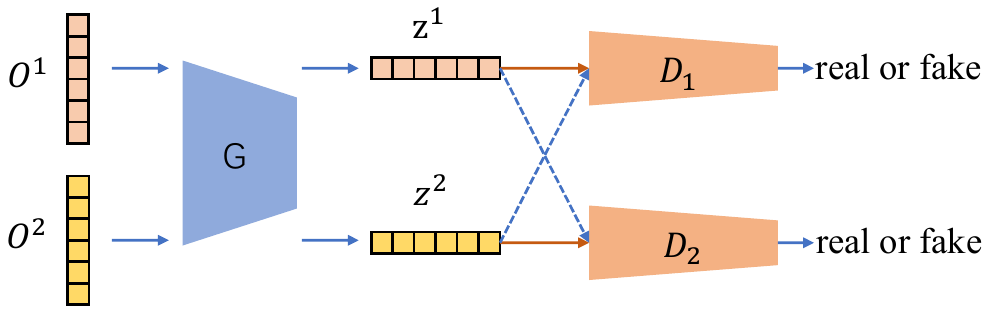}
  \caption{An example to illustrate the MDGAN.}
  \label{mdgan}
  \end{center}
\end{figure}

To better present, we take an example about how MDGAN works on two domains. As shown in Fig. \ref{mdgan}, we have two kinds of inputs, $O^1$ and $O^2$, from two domains. When feeding them into the generator $G$, we can obtain the corresponding features $z^1$ and $z^2$, respectively. After that, we feed $z^1$ and $z^2$ to the discriminator $D_1$, where $D_1$ is expected to output a ``real" label for $z^1$ and output a ``fake" label for $z^2$. By jointly training $G$ and $D_1$ based on a min-max optimization, the distribution of $z^2$ is expected to move towards that of $z^1$.  In the meantime, we also feed $z^1$ and $z^2$ to the discriminator $D_2$. Differently, $D_2$ is expected to output a ``real" label for $z^2$ and output a ``fake" label for $z^1$. By jointly training $G$ and $D_2$, the distribution of $z^1$ is expected to move towards that of $z^2$.
In this way,  $z^1$ and $z^2$ generated by the generator $G$ become domain-invariant representations.

\textcolor{black}{Formally, we define $G(O^i; \theta_g): O^i \rightarrow z$. The generator $G$ sharing with parameter $\theta_g$ across domains maps outputs $O^i$ from the $i^{th}$ domain into the latent feature $z^i$. After that, we obtain latent features $\{z^i\}_{i=1}^M$ of model outputs.
We also define $M$ discriminators $\{D^j(z^i; \theta_d^i)\}_{j=1}^{M}$.
Each discriminator $D^j(z^i): z^i \rightarrow [0,1]$ outputs a scalar representing the probability that $z^i$ comes from the $j^{th}$ domain.
The label of latent features $z^i$ is defined as \textbf{Real} for the discriminator $D^j(z^i)$ when $j=i$, while \textbf{False} when $j \neq i$.}



The training goal of $D^j$ is to maximize the log probability of assigning the correct label to features both from the $i^{th}$ domain and other domains, while the generator $G$ is trained against the discriminator to minimize the probability. In other words, it is a min-max game between the $j^{th}$ discriminator $D^j$ and generator $G$ with a value function $V$, formulated as:
\begin{equation}
\label{equ:minmax}
\begin{aligned}
\underset{G}{\textbf{min}} \; \underset{D^j}{\textbf{max}} \, V&(D^j, G) =\mathbb{E}_{x \sim O^j}[logD^j(G(x))] \\
& + \sum_{j \neq i}\mathbb{E}_{x\sim O^i}[log(1-D^j(G(x)))].
\end{aligned}
\end{equation}
During optimizing the min-max adversarial value function for $G$ and $D^j$, the generator $G$ can gradually produce latent features $z^j$ from $j^{th}$ domain, which are close to latent features from other domain. Once $G$ and all $D$ are well trained, $G$ is able to embed multi-domain model outputs into an invariant feature space, where each discriminator cannot determine which domain the outputs are from. Therefore, the latent features $\{z^i\}_{i=1}^M$ become domain-invariant features. 
\textcolor{black}{Note that our proposed MDGAN does not suffer from mode collapse. This is because mode collapse is an issue in generative tasks using GANs, where the model fails to generate diverse patterns and instead produces only a limited set of modes. In our approach, the role of generator $G$ in the MDGAN is not to generate diverse features. Instead, $G$ functions as an encoder, encoding the model's outputs from different domains into invariant features. Therefore, it does not suffer from the problem of mode collapse.}

\subsection{Domain-agnostic Reverse Meta-Model}
\label{sec:metamodel}
After obtaining domain-invariant features $\{z^i\}_{i=1}^M$, we aim to classify them as model attributes $Y$ through the domain-agnostic reverse meta-model. Consider there are $K$ types of attributes and each attribute contains $N^k$ possible values, we build $K$ domain-agnostic reverse meta-models $\Phi=\{\phi_1, \phi_2, ..., \phi_K\}$. Therefore, the probability $p_k(z)$ for the $k^{th}$ attribute is obtained by:
\begin{equation}
  p_k(z) = \textbf{softmax}(\phi_k(z)).
\end{equation}
Note that the $p_k(z)$ is a $N^k$-dimensional vector that contains $N^k$ attribute value, and each dimension represents the probability of the attribute value.



The target is to minimize the cross entropy between the predicted attributes probability $p_k(z)$ and ground-truth of model attribute values $y^k$:
\begin{equation}
  \label{equ:metace}
  \begin{aligned}
  \underset{\Phi}{\textbf{min}} \; \sum_{k=1}^K \mathbb{E}_{z \sim \{z^i\}_{i=1}^M} \left[-y_k^{T}log(p_k(z))\right].
  \end{aligned}
\end{equation}
During the inference phase, we input queries to the black-box model trained from an unknown domain to generate outputs. These outputs are then embedded as domain-invariant features by the generator $G$. Subsequently, the reverse meta-model $\Phi$ classifies these domain-invariant features, achieving domain-agnostic attributes predictions for the black-box model.
\begin{algorithm*}
  \caption{Training Strategy}
  
  \SetKw{Initialization}{Initialization}

    \KwIn{Batch size $b$, learning rate $\alpha$, $\beta$, number of attribute types $K$, multi-domain model outputs $O$, trade-off scalar $\lambda$}
    
    \KwOut{Generator $G$, meta-model $\Phi$, discriminators $\{D^j\}_{j=1}^M$}
    
    {\textbf{Initialize:} Initialize parameter $\theta_g$ of generator $G$, parameter $\theta_d^i$ of discriminators $\{D^j\}_{j=1}^M$ and parameter $\theta_c$ of domain-agnostic meta-model $\Phi$ with normal distribution} \\
   
    \While {\textcolor{black}{difference in training loss $\mathcal{L}_c$ of meta-model $\Phi$ between two consecutive epochs $ \ge \epsilon$}}{
      Random sample $b$ samples $O_b^i$ from outputs $O^i$ in each domain\\
      \For {$j = 1, ..., M$}{
        Take samples in the $j^{th}$ domain as \textbf{Real} samples $X= O^j_b = \{x^1, x^2, ..., x^b\}$ \\
        \For {$i =1,...,M \;and\;j \neq i$}{ 
            Take samples in the $j^{th}$ domain as \textbf{Fake} samples $\bar{X}_i= O^i_b = \{\bar{x}_i^1, \bar{x}_i^2, ..., \bar{x}_i^b\}$
        }
        Update the discriminator $D^j$ by gradient descent: \\
        $\theta_{d}^j := \theta_{d}^j - \alpha \nabla_{\theta_{d}^j}\left\{\sum_{k=1}^b \left[logD^j(G(x^k)\right] + \sum_{j\neq i}\left[\sum_{k=1}^blog(1-D^j(G(\bar{x}^k_i)))\right]\right\}$
      }
      
      Construct $X^* = X \cup \bar{X} = \{x^1, x^2, ..., x^{bM}\}$ and 
      $Z^* = G(X^*) = \{z^1, z^2, ..., z^{bM}\}$ \\
      Set the corresponding labels as $Y^* = \{y^1, y^2, ..., y^{bM}\}$ \\
      Calculate gradient of $\theta_c$ and $\theta_g$ by: \\
       $\bm{grad_{c}} = \nabla_{\theta_{c}}\mathcal{L}_c = \nabla_{\theta_{c}} \left\{ \sum_{k=1}^K \sum_{l=1}^{bM} \left[-{y^l_k}^Tlog(p(z^l_k))\right] \right\}$ \\
      $\bm{grad_{g}} = \nabla_{\theta_{g}, \theta_{c}}\left\{\sum_{j\neq i}\sum_{l=1}^b\left[log(1-D^j(G(\bar{x}^l_i)))\right] - \lambda\sum_{k=1}^K \sum_{l=1}^{bm}\left[{y^l_k}^Tlog(p(z^l_k))\right]\right\}$ \\
    Update the meta-model $\Phi$ and generator $G$ together: \\
      $\theta_c := \theta_c - \beta \cdot \bm{grad_{c}}$ and 
      $\theta_g := \theta_g - \alpha \cdot \bm{grad_{g}}$
    } 
  \label{alg:training}
\end{algorithm*}
\subsection{Overall Objective and Training Strategy}
\label{sec:training}
After introducing all the components, we give the final loss function based on Eq. \ref{equ:minmax} and \ref{equ:metace} as:
\begin{equation}
  \label{equ:final}
  \begin{aligned}
    \underset{G, \Phi}{\textbf{min}} \; \underset{D^j, 1 \leq j \leq M}{\textbf{max}} \, V&(D^j, G)  = \mathbb{E}_{x \sim O^j}\left[logD^j(G(x))\right] \! \\
    & + \! \sum_{j \neq i}\mathbb{E}_{x\sim O^i}\left[log(1-D^j(G(x)))\right]  \\
    & + \lambda\  \sum_{k=1}^K\mathbb{E}_{z \sim \{z^i\}_{i=1}^M} \left[-y_k^Tlog(p_k(z))\right]. 
  \end{aligned}
\end{equation}
where $\lambda$ is a trade-off parameter.

The training strategy is as follows: we first optimize all discriminators $D^i$, and then jointly optimize the generator and the domain-agnostic reverse meta-model. We repeat the above processes until the algorithm converges.
The proposed optimization strategy is presented in Algorithm \ref{alg:training}.

\section{Experiments}
\begin{table}[t]
\renewcommand\arraystretch{0.75}
\centering
\caption{Attributes and the corresponding values.}
\begin{tabular}{ll}\\
\toprule[1.3pt]
\textbf{Attribute}  & \textbf{Values} \\ \toprule [1.3pt]
\#Activation (act) & ReLU, PReLU, ELU, Tanh \\  \midrule
\#Dropout (drop) & Yes, No  \\  \midrule
\#Max pooling (pool) & Yes, No  \\  \midrule
\#Kernel size (ks) & 3, 5   \\  \midrule
\#Conv layers (conv)& 2, 3, 4  \\  \midrule
\#FC layers (fc)& 2, 3, 4   \\  \midrule
\#Optimizer (opt)& SGD, ADAM, RMSprop   \\  \midrule
\#Batch size (bs)& 32, 64, 128  \\ \midrule
\#Batchnorm (bn) & Yes, No   \\  \midrule[1.3pt]
\end{tabular}
\label{table:attr}
\end{table}

\label{others}

\subsection{Dataset Construction}
\label{subsec:dataset_construction}
Following \cite{oh2018towards2}, we construct the modelset by training models that enumerate all possible attribute values. The details of the attributes and their values are shown in Table \ref{table:attr}. There are a total of $K=9$ types of attributes for each model in the modelsets, which adheres the following scheme: $N_c=\{2,3,4\}$ convolution layers, $N_f=\{2,3,4\}$ fully-connected layers \footnote{\textcolor{black}{Not that a model with more layers may be mistakenly classified as an overtrained model. However, this issue do not exist in practical applications, because we did not devise an attribute to determine if a model is overtrained. In practical applications, people tend to deploy models with strong generalization capabilities rather than overtrained ones. Therefore, to ensure accurate model reverse engineering, the white-box models we train are not overtrained either.
We select the model that performs best on the validation set as the final white-box model, thereby ensuring its generalization performance.}}. Each convolution layer contains a $k \times k$ kernel ($k=\{3, 5\}$), an optional batch normalization, an optional max-pooling, and a non-linear activation function in sequence. Each fully connected layer consists of a linear transformation, a non-linear activation, and an optional dropout in sequence. We set the dropout ratio to $0.1$ in our experiments. When training white-box models, optimizers are selected from $\{$SGD, ADAM, RMSprop$\}$ with a batch size $32$, $64$ or $128$, respectively. \textcolor{black}{The statistics of PACS modelset and MEDU modelset are listed in the Appendix.B}
By enumerating all possible model attributes, a total of $5,184$ distinct models can be obtained. In addition, we initialize each kind of model with random seeds from 0 and 999, yielding 5,184,000 unique models. 

We construct PACS modelset and MEDU modelset to evaluate our method. The details are as follows:

\textbf{PACS modelset} comprises a set of models trained on the PACS dataset. The PACS dataset is an image dataset that has been widely used for OOD generalization \cite{li2017deeper}. 
We utilize 3 domains, including Photo (1,670 images), Cartoon (2,344 images), and Sketch (3,929 images), and each domain contains 7 kind of classes. For each modelset domain, we randomly sample and train 5,000, 1,000, and 1,000 from 5,184,000 white-box models as the training, validation, and testing models, respectively.

\textbf{MEDU modelset} consists of a set of models trained on MEDU dataset. MEDU is a hand-written digit recognition dataset, with 4 domains collected from MNIST \cite{lecun_mnist_1998}, USPS \cite{hull_database_1994}, DIDA \cite{DIDA2020} and EMNIST \cite{cohen_emnist_2017}. Each domain contains different styles of hand-written digits from 0 to 9. Similar to PACS modelset, for each modelset domain, we randomly sample and train 5,000, 1,000, and 1,000 from 5,184,000 white-box models as the training, validation, and testing models.


\subsection{Implementation Details of DREAM}
\label{app:impl_details}
In the experiment, we set the number of queries $N$ to 100. 
We use Adam \cite{kingma2014adam} as the optimizer, where the learning rate $\alpha$  is set to $10^{-5}$  for the generator and discriminators, and the learning rate $\beta$ is set to $10^{-4}$ for the domain-agnostic meta-model. 
The batch size $b$ is set as 100. 
The trade-off parameter $\lambda$ is tuned from $\{0.001, 0.01, 0.1, 1, 10\}$ based on the validation set. 

In addition, the MDGAN is composed of a generator and multiple discriminators. The generator consists of two linear layers with ReLU activation. The dimension of the input layer of the generator is determined by the query number $N$ and class category number $C$.  As for the experiment conducted on MEDU modelset, the input dimension is $NC = 1000$ (the query number is $N=100$, and the number of classes if $C=10$). In the case of the PACS modelset, the input dimension is 700 ($N=100$, $C=7$). The output dimensions of the subsequent two linear layers are 500 and 128, respectively. Each discriminator consists of three linear layers. The first two layers employ ReLU as the activation function, while the last layer utilizes the Sigmoid activation function. The output dimensions of discriminator layers are 512, 256, and 1 respectively. Additionally, we implement the domain agnostic meta-models as $K=9$ MLPs. Each MLP consists of 3 layers, and it takes latent features produced by generator $G$ as input. The output dimensions of the MLP are 128, 64 and $N^k$. All experiments are conducted on 4 NVIDIA RTX 3090 GPUs, PyTorch 1.11.0 platform \cite{paszke2019pytorch}.

\subsection{Baselines}
We compare our DREAM with 7 baselines including Random choice, SVM, KENNEN \cite{oh2018towards2}, SelfReg \cite{kim2021selfreg}, MixStyle \cite{zhou2021mixstyle}, MMD \cite{li_domain_2018} and SD \cite{pezeshki2021gradient}. Additionally, we take four typical OOD generalization methods—SelfReg, MixStyle, MMD, and SD—as baselines to validate the effectiveness of our framework in learning domain-invariant features. 

The details of the baseline methods are presented as follows:

\textbf{SVM.} The SVM serves as a baseline, representing a non-deep learning method. We directly input the multi-domain probability outputs into SVM classifiers to predict attributes. 
\textcolor{black}{We adopt the One-vs-One (OvO) strategy. Specifically, for an attribute with $N^k$ possible values, we trained a binary SVM for each pair of attribute values, resulting in a total of $N^k(N^k-1)/2$ SVMs. When classifying a new sample, we apply all trained SVM classifiers and use a voting mechanism for each attribute value. The attribute value receiving the most votes is selected as the final prediction. For 2-valued attributes, we only need to train a single binary SVM.}

\textbf{KENNEN*.} We employ a variant of KENNEN (denoted as KENNEN* ). It takes fixed queries as input, which is the same as our approach. We embed the multi-domain probability outputs into feature space and use MLPs to predict attributes. Similar to SVM, KENNEN does not differentiate between outputs from different domains. In addition, to ensure a fair comparison, the network structure keeps consistent with the domain-agnostic reverse meta-model in DREAM.

\textbf{SelfReg, MixStyle, MMD and SD.} The approach taken by SelfReg to learn invariant features involves pulling samples of similar categories between all domains closer together while pushing samples of different categories further apart. The motivation behind MixStyle is based on the observation that the styles of domain images share significant similarities. In particular, MixStyle captures style information through the final layer and performs style mixing at that layer. The MMD adopts maximum mean discrepancy loss between each two domains. The SD proposes the Spectral Decoupling method to relieve the gradient starvation phenomenon during the training of the network, to boost the performance of OOD generalization. We embed the multi-domain probability outputs into feature space and apply the OOD generalization method, SelfReg, MixStyle, MMD, and SD, respectively to learn invariant features. Then the learned invariant features are fed into MLPs for predicting attributes. For a fair comparison, the network structure of MLPs is consistent with the domain-agnostic reverse meta-model in DREAM.

We adopt the \textbf{leave-one-domain-out} strategy to split the source and target domains for PACS and MEDU modelsets. For each modelset, we take one domain as the target domain and the rest as source domains in turn. The experiment is run for 10 trials, and the average results are reported. 


\begin{table*}[ht]
  \renewcommand\arraystretch{0.95}
  \caption{Attribute classification acc. (\%) on PACS modelset. \textcolor{red}{\textbf{Red}} and \textcolor{black}{\underline{blue}}  indicate the best and second best performance, respectively.}
  \vspace{-0.1in}
  \label{table:result_PACS}
  \centering
  \setlength\tabcolsep{5pt}
  \begin{tabular}{c|cccccccccccc}
  \toprule[1.3pt]
  \multirow{3}{*}{\makecell[c]{Target\\Domain}} & \multirow{2}{*}{Method} & \multicolumn{9}{c}{Attributes} & \multirow{2}{*}{Avg}  \\
  \cline{3-11}
   & & \#act & \#drop & \#pool & \#ks & \#conv & \#fc & \#opt& \#bs & \#bn \\
  \cline{2-12}
  &Random &25.00 &50.00 &50.00 &50.00 &33.33&33.33 &33.33 &33.33 &50.00 &39.81 \\
  \hline
  \multirow{8}{*}{Photo}&SVM &37.80 &50.30 &54.80 &53.60 &34.00 &36.60 &37.00 &\color{blue}\underline{45.70} &58.80 &45.40 \\
  &KENNEN* &39.07 &50.68 &59.42 &61.31&\color{blue}\underline{36.18} &39.33 &37.88 &44.16 &59.74 &47.53\\
  &SelfReg &25.58 &52.26 &54.98 &50.18 &34.12 &35.25 &34.61 &33.78 &50.76 &41.28 \\
  &MixStyle &\color{blue}\underline{39.63} &53.23 &\color{blue}\underline{61.83} &59.44 &35.66 &38.75 &37.89 &43.75 &57.09 &47.47 \\
  &MMD &38.88 &\color{blue}\underline{54.70} &60.46 &56.54 &35.38 &36.66 &35.66 &40.50 &61.04 &46.65 \\
  
  &SD &38.70 &51.06 &58.86 &\color{blue}\underline{62.21} &35.84 &\color{blue}\underline{40.05} &\color{blue}\underline{39.23} &44.34 &\color{blue}\underline{62.12} &\color{blue}\underline{48.04} \\
  &\textbf{DREAM} &\color{red}\textbf{43.84} &\color{red}\textbf{59.19} &\color{red}\textbf{66.09} &\color{red}\textbf{64.24} &\color{red}\textbf{39.59} &\color{red}\textbf{42.04} & \color{red}\textbf{40.49} &\color{red}\textbf{47.83} &\color{red}\textbf{68.12} &\color{red}\textbf{52.38} \\
  \hline
  \multirow{8}{*}{Cartoon} &SVM &25.80 &49.20 &50.70 &55.80 &\color{blue}\underline{37.20} &38.10 &30.80 &\color{blue}\underline{42.30} & \color{blue}\underline{65.30} &43.91 \\
  &KENNEN* &32.99 &52.50 &54.23 &56.57 &37.19 &\color{blue}\underline{40.53} &33.47 &37.17 &\color{red}\textbf{68.39} &45.89\\
  &SelfReg &25.97 &51.42 &\color{blue}\underline{56.20} &50.03 &35.04 &35.52 &\color{blue}\underline{36.09} &35.58 &56.17 &42.44 \\
  &MixStyle &32.10 &50.76 &55.44 &54.18 &36.18 &37.87 &34.65 &38.69 &60.26 &44.46 \\
  &MMD &29.56 &53.02 &54.70 &53.82 &35.38 &36.36 &35.98 &37.24 &57.58 &43.75 \\
  
  &SD &\color{blue}\underline{33.52} &\color{blue}\underline{54.06} &54.12 &\color{blue}\underline{56.69} &36.84 &\color{red}\textbf{41.02} &35.61 &36.12 &65.12 &\color{blue}\underline{45.90} \\
  
  &\textbf{DREAM} &\color{red}\textbf{37.53} &\color{red}\textbf{55.89} &\color{red}\textbf{61.18} &\color{red}\textbf{57.32} &\color{red}\textbf{38.58} &39.60 &\color{red}\textbf{38.32} &\color{red}\textbf{45.01} &65.16 &\color{red}\textbf{48.73} \\
  \hline
  
  \multirow{8}{*}{Sketch}&SVM &23.80 &47.60 &47.40 &45.80 &33.80 &34.50 &31.80 &34.30 &53.10 &39.12 \\
  &KENNEN* &34.64 &50.10 &53.07 &52.01 &34.61 &37.11 &35.78 &\color{blue}\underline{37.04} &55.27 &43.29\\
  &SelfReg &27.07 &\color{blue}\underline{54.32} &51.39 &53.07 &36.99 &36.82 &35.47 &34.17 &\color{blue}\underline{61.80} &43.46 \\
  &MixStyle &\color{blue}\underline{37.78} &51.71 &54.16 &\color{blue}\underline{53.60} &34.53 &36.16 &\color{blue}\underline{36.36} &36.02 &59.42 &44.42 \\
  &MMD &31.96 &52.94 &56.84 &52.78 &\color{blue}\underline{38.18} &\color{blue}\underline{38.20} &36.20 &35.92 &57.56 &\color{blue}\underline{44.51} \\

  &SD &34.82 &52.51 &\color{blue}\underline{56.89} &51.21 &34.23 &38.12 &35.91 &36.72 &54.23 &43.85 \\
  
  &\textbf{DREAM}
  &\color{red}\textbf{39.71} &\color{red}\textbf{57.74} &\color{red}\textbf{64.73} &\color{red}\textbf{60.79} &\color{red}\textbf{40.79} &\color{red}\textbf{40.14} &\color{red}\textbf{43.54} &\color{red}\textbf{43.80} &\color{red}\textbf{72.51} &\color{red}\textbf{51.53} \\
  \bottomrule[1.3pt]
\end{tabular}
\vspace{-0.1in}
\end{table*}

\begin{table*}[ht]
\renewcommand\arraystretch{0.95}
  \caption{Attribute classification acc. (\%) on MEDU modelset. \textcolor{red}{\textbf{Red}} and \textcolor{black}{\underline{blue}}  indicate the best and second best performance, respectively.}
  \vspace{-0.1in}
  \label{table:result_MEDU}
  \centering
  \setlength\tabcolsep{5pt}
  \begin{tabular}{c|cccccccccccc}
  \toprule[1.3pt]
  \multirow{3}{*}{\makecell[c]{Target\\Domain}} & \multirow{2}{*}{Method} & \multicolumn{9}{c}{Attributes} & \multirow{2}{*}{Avg}  \\
  \cline{3-11}
   & & \#act & \#drop & \#pool & \#ks & \#conv & \#fc & \#opt& \#bs & \#bn \\
   \cline{2-12}
  &Random &25.00 &50.00 &50.00 &50.00 &33.33&33.33 &33.33 &33.33 &50.00 &39.81 \\
  \cline{1-12}
  \multirow{8}{*}{MNIST} &SVM &45.60 &49.40 &62.90 &\color{red}\textbf{59.20} &38.80 &\color{red}\textbf{40.10} &35.50 &35.00 &75.30 &49.09 \\
  &KENNEN* &\color{red}\textbf{51.18} &50.67 &62.99 &57.36 &38.32 &35.84 &41.57 &35.75 &77.87 &50.17\\
  &SelfReg &28.00 &53.57 &53.43 &50.78 &35.97 &36.39 &35.98 &36.23 &53.96 &42.70 \\
  &MixStyle &50.27 &51.72 &62.66 &57.32 &37.88 &36.34 &\color{blue}\underline{43.11} &\color{blue}\underline{38.00} &\color{red}\textbf{82.61} &51.10 \\
  &MMD &44.57 &\color{blue}\underline{59.67} &\color{red}\textbf{66.37} &57.27 &\color{blue}\underline{39.63} &37.27 &42.10 &37.60 &81.37 &\color{blue}\underline{51.76} \\
  & SD & 49.60 & 49.40& 62.40& 52.30& 37.10& 36.70& 38.90& 35.30& 81.50& 49.24\\
  
  &\textbf{DREAM} &\color{blue}\underline{51.01} &\color{red}\textbf{62.32} &\color{blue}\underline{64.28} &\color{blue}\underline{58.39} &\color{red}\textbf{40.96} &\color{blue}\underline{38.11} &\color{red}\textbf{45.37} &\color{red}\textbf{38.96} &\color{blue}\underline{81.99} &\color{red}\textbf{53.49} \\
  \hline
  \multirow{8}{*}{EMNIST}&SVM &40.00 &48.70 &\color{blue}\underline{69.20} &51.60 &40.20 &36.90 &35.80 &30.10 &79.90 &48.04 \\
  &KENNEN* &\color{red}\textbf{45.66} &51.01 &65.26 &53.25 &40.28 &36.35 &41.96 &36.16 &81.30 &50.14\\
  &SelfReg &27.29 &52.83 &53.32 &52.85 &33.68 &35.05 &35.26 &35.32 &53.74 &42.15 \\
  &MixStyle &43.68 &51.35 &67.87 &57.15 &42.50 &\color{blue}\underline{39.30} &\color{blue}\underline{42.10} &38.79 &82.46 &51.69 \\
  &MMD &42.03 &\color{blue}\underline{58.43} &66.27 &\color{red}\textbf{60.80} &40.80 &38.67 &40.00 &\color{blue}\underline{39.97} &84.00 &\color{blue}\underline{52.33} \\
  & SD & 43.60& 48.90& 59.20& 60.10& \color{red}\textbf{44.50}& 35.10& 43.60& 33.80& \color{blue}\underline{88.80}& 50.84\\
  &\textbf{DREAM} &\color{blue}\underline{45.55} &\color{red}\textbf{64.98} &\color{red}\textbf{74.16} &\color{blue}\underline{60.71} &44.45&\color{red}\textbf{42.45} &\color{red}\textbf{47.37} &\color{red}\textbf{41.03} &\color{red}\textbf{91.00} &\color{red}\textbf{56.86} \\
  \hline
  \multirow{8}{*}{DIDA}&SVM &45.00 &47.80 &54.60 &45.50 &29.40 &37.60 &\color{red}\textbf{43.30} &36.50 &\color{red}\textbf{63.70} &44.82 \\
  &KENNEN* &42.73 &52.06 &55.27 &52.02 &34.89 & 38.90 &38.98 &36.27 &54.97 &45.12\\
  &SelfReg &26.31 &54.29 &53.23 &52.33 &34.96 &35.72 &36.49 &35.39 &59.11 &43.09 \\
  &MixStyle &\color{blue}\underline{45.26} &52.32 &55.91 &51.39 &34.22 &38.70 &38.31 &\color{blue}\underline{38.03} &57.44 &45.73 \\
  &MMD &39.00 &\color{blue}\underline{59.20} &\color{red}\textbf{59.63} &\color{blue}\underline{55.93} &\color{blue}\underline{35.93} &38.33 &37.93 &37.50 &54.40 &46.43 \\
  & SD & 46.50& 51.80& 52.70& 51.90& 34.90& \color{red}\textbf{45.30}& \color{blue}\underline{42.70}& 36.80& 59.50&  \color{blue}\underline{46.92}\\
  &\textbf{DREAM} &\color{red}\textbf{49.63} &\color{red}\textbf{64.50} &\color{blue}\underline{59.30} &\color{red}\textbf{57.13} &\color{red}\textbf{39.52} &\color{blue}\underline{44.59} &42.09 &\color{red}\textbf{40.19} &\color{blue}\underline{59.68} &\color{red}\textbf{50.74} \\
  \hline

  \multirow{8}{*}{USPS}&SVM &\color{red}\textbf{43.40} &50.50 &47.60 &52.50 &30.30 &32.30 &\color{red}\textbf{41.00} &36.60 &49.40 &42.62 \\
  &KENNEN* &\color{blue}\underline{43.38} &50.88 &51.41 &53.19 &36.35&35.59 &36.66 &34.56 &55.62 &44.18\\
  &SelfReg &26.81 &52.16 &55.46 &52.47 &36.18 &\color{blue}\underline{36.43} &36.53 &35.90 &55.34 &43.03 \\
  &MixStyle &41.05 &53.80 &50.49 &52.93 &35.26 &33.68 &36.92 &34.75 &59.34 &44.25 \\
  &MMD &39.33 &\color{blue}\underline{55.87} &52.67 &\color{blue}\underline{53.23} &\color{red}\textbf{39.20} &34.33 &35.90 &\color{blue}\underline{36.90} &60.73 &\color{blue}\underline{45.35} \\
  & SD & 41.90& 50.40& \color{blue}\underline{58.00}& 52.30& 33.10& 35.30& 36.60& 34.30 & \color{blue}\underline{61.60}& 44.83 \\
  &\textbf{DREAM} &42.34 &\color{red}\textbf{58.72} &\color{red}\textbf{58.58} &\color{red}\textbf{54.41} &\color{blue}\underline{37.90} &\color{red}\textbf{37.81} &\color{blue}\underline{40.42} &\color{red}\textbf{38.36} &\color{red}\textbf{63.39} &\color{red}\textbf{47.99} \\
  
  \bottomrule[1.3pt]
\end{tabular}
\vspace{-0.1in}
\end{table*}

\begin{table*}[t]
  \caption{Accuracy of model extraction using different extraction model structures. The extraction model structures are "same to the victim", "random", or "inferred by DREAM".
}
  \label{table:me}
  \centering
  \setlength\tabcolsep{5pt}
  \begin{threeparttable}
  \begin{tabular}{ccccc}
  \toprule[1.3pt]
  \multirow{2}{*}{Dataset} & 
  \multirow{2}{*}{Victim Model} &
  \multicolumn{3}{c}{Architecture of the Extraction Model} \\  
  \cline{3-5}
   & & \#Same to Victim & \#Random & \#Inferred by DREAM  \\
   \hline
 
  \multicolumn{1}{c}{MNIST} &$86.43\%_{(1.00\times)}$ & $68.46\%_{(0.79\times)}$
  &$45.88\%_{(0.53\times)}$
  &$62.81\%_{(0.73\times)}$  \\
  \bottomrule[1.3pt]
\end{tabular}
\end{threeparttable}
\end{table*}
\begin{figure*}[t]
	\centering
	\setcounter {subfigure} {0}{
    	\begin{minipage}{0.22\linewidth}
    		\centering
    		\includegraphics[width=1\linewidth]{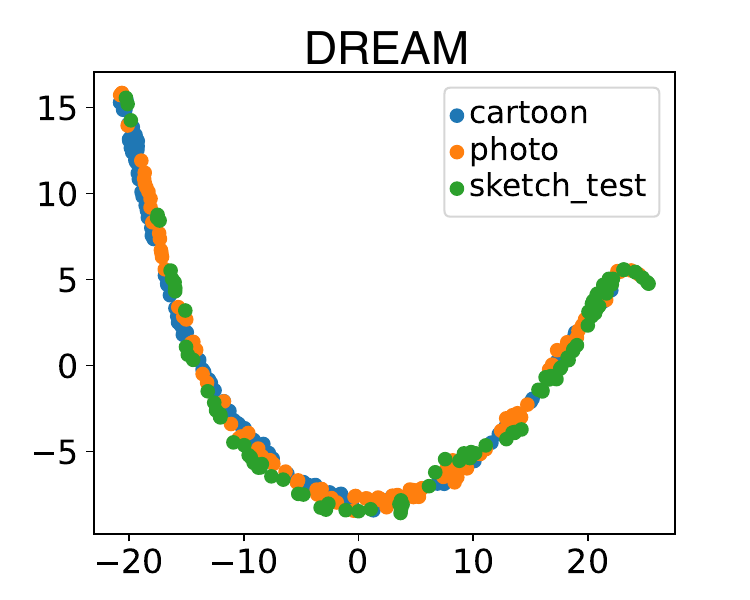}
    		\label{fig:param_id2_S}
    	\end{minipage}
    	\begin{minipage}{0.22\linewidth}
    		\centering
    		\includegraphics[width=1\linewidth]{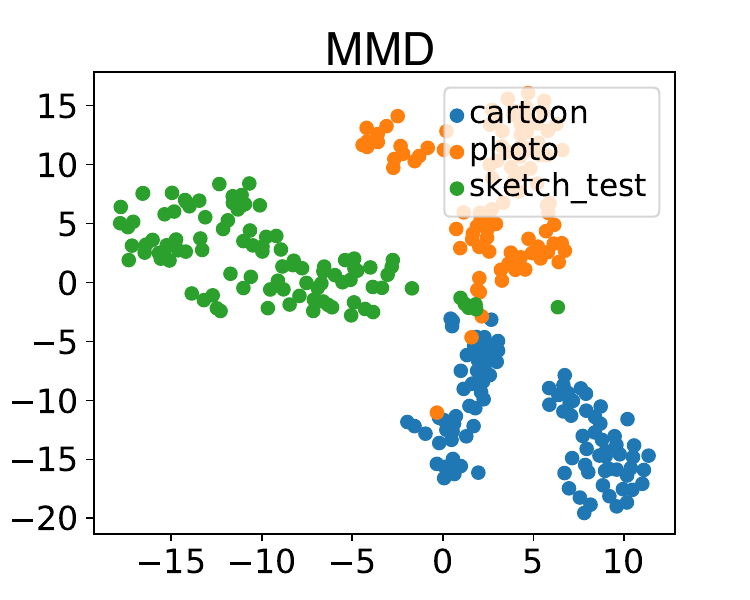}
    		\label{fig:param_id2_P}
    	\end{minipage}
    	\begin{minipage}{0.22\linewidth}
    		\centering
    		\includegraphics[width=1\linewidth]{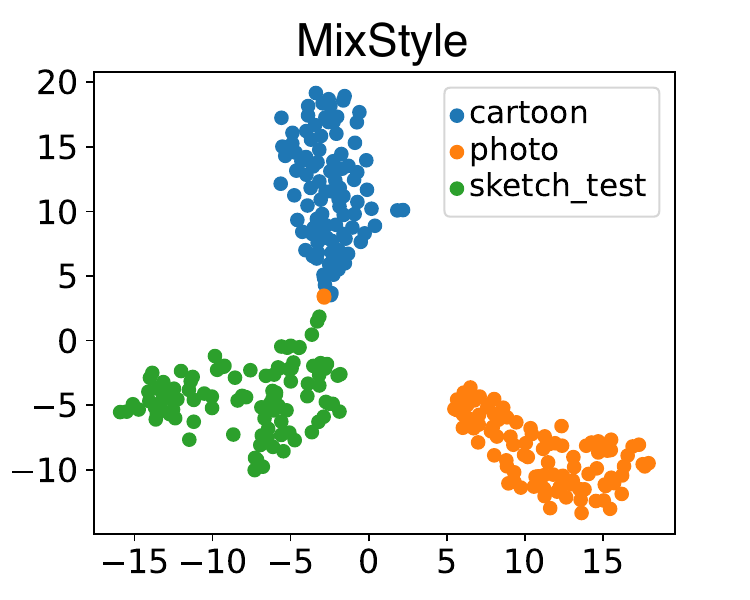}
    		\label{fig:param_id2_C}
    	\end{minipage}
    	\begin{minipage}{0.22\linewidth}
    		\centering
    		\includegraphics[width=1\linewidth]{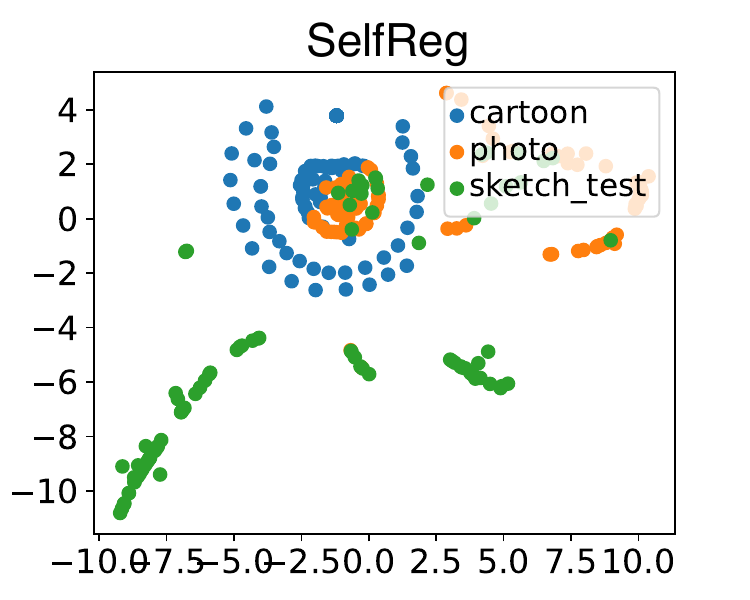}
    		\label{fig:param_id2_S}
    	\end{minipage}
	}
	\caption{T-SNE visualization of features of different domains produced by DREAM, MMD, MisStyle and SelfReg on PACS modelset.}
	\label{fig:tsne_DREAM} 
	\vspace{-0.1in}
\end{figure*}

\begin{table*}[t]
  \caption{Model attribute classification accuracy (\%) on S of PACS modelset. \textcolor{red}{\textbf{Red}} and \textcolor{black}{\underline{blue}}  indicate the best and second best performance, respectively. \textbf{DREAM*} is the result of domain shift
    scenario, trained with only five classes (except dog and elephant), while the black-box model is trained by whole seven classes in PACS dataset.}
  \label{table:ccs}
  \centering
  \setlength\tabcolsep{5pt}
  \begin{threeparttable}
  \begin{tabular}{c|cccccccccccc}
  \toprule[1.3pt]
  \multirow{3}{*}{\makecell[c]{Target\\Domain}} & \multirow{2}{*}{Method} & \multicolumn{9}{c}{Attributes} & \multirow{2}{*}{Avg}  \\
  \cline{3-11}
   & & \#act & \#drop & \#pool & \#ks & \#conv & \#fc & \#opt& \#bs & \#bn \\
   \cline{2-12}
  &Random &25.00 &50.00 &50.00 &50.00 &33.33&33.33 &33.33 &33.33 &50.00 &39.81 \\
  \hline
  \multirow{8}{*}{Sketch}&SVM &23.80 &47.60 &47.40 &45.80 &33.80 &34.50 &31.80 &34.30 &53.10 &39.12 \\
  &KENNEN* &34.64 &50.10 &53.07 &52.01 &34.61 &37.11 &35.78 &37.04 &55.27 &43.29\\
  &SelfReg &27.07 &54.32 &51.39 &53.07 &36.99 &36.82 &35.47 &34.17 &61.80 &43.46 \\
  &MixStyle &37.78 &51.71 &54.16 &53.60 &34.53 &36.16 &36.36 &36.02 &59.42 &44.42 \\
  &MMD &31.96 &52.94 &56.84 &52.78 &38.18 &38.20 &36.20 &35.92 &57.56 &44.51 \\
  &SD &34.82 &52.51 &56.89 &51.21 &34.23 &38.12 &35.91 &36.72 &54.23 &43.85 \\
  &\textbf{DREAM} 
  &\color{blue}\underline{39.71} &\color{red}\textbf{57.74} &\color{red}\textbf{64.73} &\color{red}\textbf{60.79} &\color{red}\textbf{40.79} &\color{red}\textbf{40.14} &\color{red}\textbf{43.54} &\color{red}\textbf{43.80} &\color{blue}\underline{72.51} &\color{red}\textbf{51.53} \\

  &\textbf{DREAM*}
&\color{red}\textbf{42.24} &\color{blue}\underline{55.68} &\color{blue}\underline{61.82} &\color{blue}\underline{58.34} &\color{blue}\underline{39.55} &\color{blue}\underline{38.39} &\color{blue}\underline{38.51} &\color{blue}\underline{41.39} &\color{red}\textbf{74.39} &\color{blue}\underline{50.03} \\
  \bottomrule[1.3pt]
\end{tabular}
\end{threeparttable}
\end{table*}
\begin{table*}
  \renewcommand\arraystretch{0.95}
  \caption{Attribute inference accuracy (\%) of reverse engineering. The source domains are outputs of Photo and Sketch model. The target domain is outputs of CIFAR model. \textcolor{red}{\textbf{Red}} and \textcolor{black}{\underline{blue}}  indicate the best and second best performance, respectively.}
  \vspace{-0.1in}
  \label{table:quite_diff_CIFAR}
  \centering
  \setlength\tabcolsep{5pt}
  \begin{tabular}{c|cccccccccccc}
  \toprule[1.3pt]
  \multirow{3}{*}{\makecell[c]{Target\\Domain}} & \multirow{2}{*}{Method} & \multicolumn{9}{c}{Attributes} & \multirow{2}{*}{Avg}  \\
  \cline{3-11}
   & & \#act & \#drop & \#pool & \#ks & \#conv & \#fc & \#opt& \#bs & \#bn \\
  \cline{2-12}
  &Random &25.00 &50.00 &50.00 &50.00 &33.33&33.33 &33.33 &33.33 &50.00 &39.81 \\
  \hline
  \multirow{8}{*}{CIFAR}&SVM & 33.91  & 49.85  & 51.87  & 50.45  & 36.43  & 36.23  & 33.80  & 41.78  & 58.73  & 43.67   \\
  &KENNEN* & 38.65  & 56.71  & 56.21  & 56.71  & 36.73  & 39.35  & 41.68  & 44.30  & 63.57  & 48.21 \\
  &SelfReg & 39.46  & 52.17  & \color{blue}\underline{60.44}  & 55.60  & \color{blue}\underline{38.65}  & 40.46  & 47.12  & 47.53  & 61.45  & 49.21 \\
  &MixStyle & \color{blue}\underline{41.78}  & 53.48  & 57.52  & 53.18  & 36.93  & 41.68  & 43.59  & 46.72  & 63.98  & 48.76  \\
  &MMD & 40.46  & \color{blue}\underline{60.54}  & 59.54  & \color{blue}\underline{57.32}  & 37.94  & \color{blue}\underline{43.49}  & \color{red}\textbf{48.84}  & \color{blue}\underline{50.25}  & 67.00  & \color{blue}\underline{51.71} \\
  &SD & 39.05  & 53.78  & 57.92  & 56.51  & 36.23  & 41.78  & 41.47  & \color{red}\textbf{50.76}  & \color{blue}\underline{67.31}  & 49.42\\
  &\textbf{DREAM} &\color{red}\textbf{45.11} &\color{red}\textbf{62.36} &\color{red}\textbf{61.35} &\color{red}\textbf{61.76} &\color{red}\textbf{41.88} &\color{red}\textbf{45.41} & \color{blue}\underline{47.53} &49.85 &\color{red}\textbf{75.88} &\color{red}\textbf{54.57} \\
  \bottomrule[1.3pt]
\end{tabular}
\vspace{-0.1in}
\end{table*}

\subsection{Comparison with Baselines}
\label{subsec:exp_result}
Table \ref{table:result_PACS} and \ref{table:result_MEDU} report the overall performance of different methods on the PACS modelset and MEDU modelset, respectively.
The leftmost column in each table indicates the target domain (the rest ones are source domains).  \textcolor{black}{In this experiment, we adopt the complete overlap setting, where the label space of the source domain and target domain is identical.}
The performance achieved by our proposed DREAM is better than that of all baselines in terms of the average accuracy of model attributes.
For individual attribute, our method outperforms other methods in most cases.
DREAM demonstrates superior performance compared to KENNEN and SVM. This suggests that, in situations where the training dataset of the target black-box model is unknown, utilizing our proposed MDGAN for learning invariant features, instead of directly inputting the outputs from the source domain into the meta-model for attribute prediction, results in enhanced generalization. DREAM also outperforms four OOD generalization methods, which highlights the superior capability of the proposed MDGAN in learning domain-invariant features from probability outputs compared to other baselines.
\subsection{Application Value of Model Reverse Engineering}
To show the application value of our method, we conduct a model extraction experiment. We adopt a popular model extraction method MAZE \cite{kariyappa_maze_2021}, where model structures are "\textbf{Same to the victim}", "\textbf{Random}", or "\textbf{Inferred by DREAM}". 
As shown in Table \ref{table:me}, The performance of model extraction using the architecture "Inferred by DREAM" significantly outperforms that achieved with a random architecture ($62.81\%$ vs. $45.88\%$). Moreover, its performance closely approaches that obtained when employing the "Same as the victim" architecture for model extraction ($62.81\%$ vs. $68.46\%$).
The rationale behind above results lies in the fact that extraction performance improves when the structure of the extracted model closely aligns with that of the victim, as discussed in \cite{sha2023can}. The proximity of the extracted model to the victim model in terms of structure and complexity enhances the likelihood of capturing identical patterns and information present in the victim model.



\subsection{Visualization of Generated Feature Space}
\textcolor{black}{
To further verify the effectiveness of our method, we utilize t-SNE \cite{van2008visualizing} to visualize samples in the feature space learned by the generator $G$ of our framework. 
The visualization is carried out on PACS modelset. We take C (cartoon) and P (photo) as source domains to train white-box models, and use S (sketch) as the unseen target domain to the train black-box model. \textcolor{black}{In this experiment, we adopt the complete overlap setting.} As shown in Fig. \ref{fig:tsne_DREAM}, we observe the features from sources and target domain are spread in the feature space for MMD, MixStyle and SelfReg method. However,
DREAM can embed samples from both the source domains and target domain into an invariant feature space, which demonstrates the superiority of DREAM.
}

\subsection{Experiment on Domain Shift Scenario}
1) We conduct class shift experiment on PACS modelset. \textcolor{black}{In this experiment, we exclude the "dog" and "elephant" classes among 7 classes for each source domain while constructing white-box models, while the target domain remain all 7 classes.}
As shown in Table \ref{table:ccs}, the average accuracy of DREAM* reaches 50.03\%. Although DREAM* experiences a slight decrease in accuracy compared to DREAM due to domain shift, it still outperforms other baselines.

\begin{table*}[t]
  \caption{Model attribute classification accuracy (\%) on P of PACS modelset using different training and testing attribute combinations. \textcolor{red}{\textbf{Red}} and \textcolor{black}{\underline{blue}}  indicate the best and second best performance, respectively.}
  \label{table:diff_arc}
  \centering
  \setlength\tabcolsep{5pt}
  \begin{threeparttable}
  \begin{tabular}{c|cccccccccccc}
  \toprule[1.3pt]
  \multirow{3}{*}{\makecell[c]{Target\\Domain}} & \multirow{2}{*}{Method} & \multicolumn{9}{c}{Attributes} & \multirow{2}{*}{Avg}  \\
  \cline{3-11}
   & & \#act & \#drop & \#pool & \#ks & \#conv & \#fc & \#opt& \#bs & \#bn \\
  \cline{2-12}
  &Random &25.00 &50.00 &50.00 &50.00 &33.33&33.33 &33.33 &33.33 &50.00 &39.81 \\
  \hline
  \multirow{8}{*}{Photo}&SVM &34.20 &51.70 &48.50 &56.10 &35.70 &36.50 &37.60 &40.50 &64.60 &45.04 \\
  &KENNEN* &37.36 &53.12 &57.79 &59.66 &38.94 &35.93 &37.92 &41.71 &63.91 &47.37\\
  &SelfReg &26.08 &52.35 &53.89 &52.70 &35.11 &33.84 &37.46 &36.42 &50.99 &42.09 \\
  &MixStyle &35.98 &54.31 &57.35 &57.43 &37.14 &35.51 &39.31 &42.07 &57.84 &46.33 \\
  &MMD &38.67 &57.16 &61.49 &58.73 &\color{blue}\underline{40.65} &39.14 &38.69 &41.06 &\color{blue}\underline{71.48} &49.67 \\
  &SD &38.70 &51.06 &58.86 &\color{blue}\underline{62.21} &35.84 &40.05 &39.23 &\color{blue}\underline{44.34} &62.12 &48.04 \\
  &\textbf{DREAM} &\color{red}\textbf{43.84} &\color{red}\textbf{59.19} &\color{red}\textbf{66.09} &\color{red}\textbf{64.24} &39.59 &\color{red}\textbf{42.04} & \color{blue}\underline{40.49} &\color{red}\textbf{47.83} &68.12 &\color{red}\textbf{52.38} \\

  &\textbf{DREAM**} &\color{blue}\underline{39.68} &\color{blue}\underline{57.61} &\color{blue}\underline{64.48} &60.79&\color{red}\textbf{40.78} &\color{blue}\underline{40.10} & \color{red}\textbf{43.54} &43.80&\color{red}\textbf{72.42} &\color{blue}\underline{51.47} \\
  \midrule[1.3pt]
\end{tabular}
\end{threeparttable}
\end{table*}

\begin{figure*}
	\centering
	\begin{minipage}{0.3\linewidth}
		\centering
		\includegraphics[width=1\linewidth]{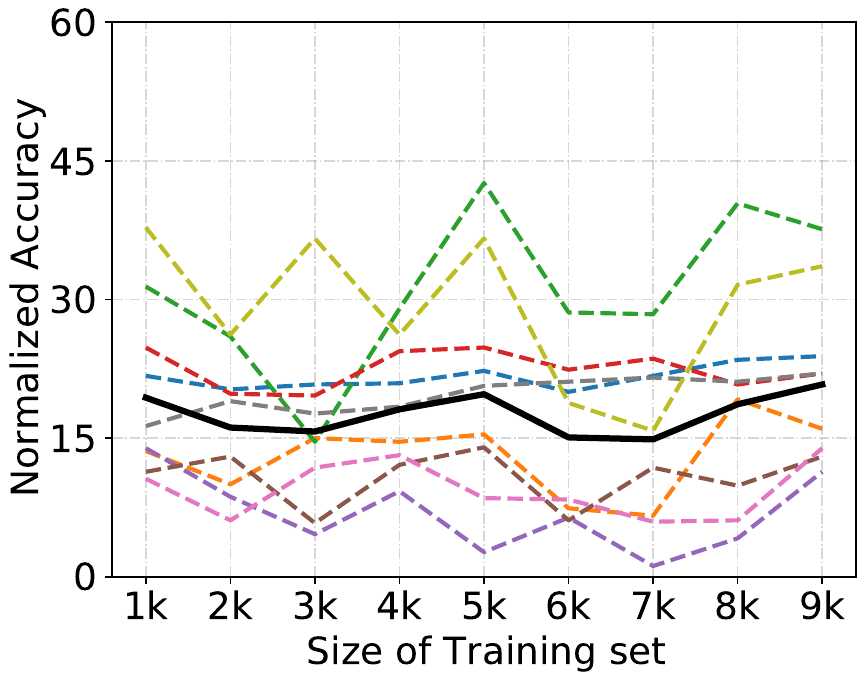}
		\label{fig:param_id2_P}
	\end{minipage}
	\begin{minipage}{0.3\linewidth}
		\centering
		\includegraphics[width=1\linewidth]{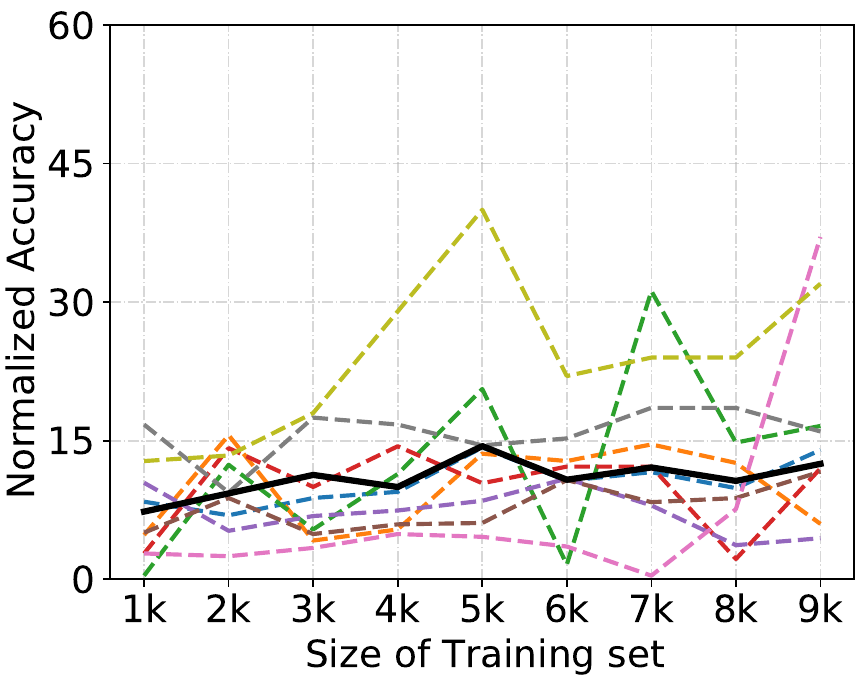}
		\label{fig:param_id3_C}
	\end{minipage}
	\begin{minipage}{0.3\linewidth}
		\centering
		\includegraphics[width=1\linewidth]{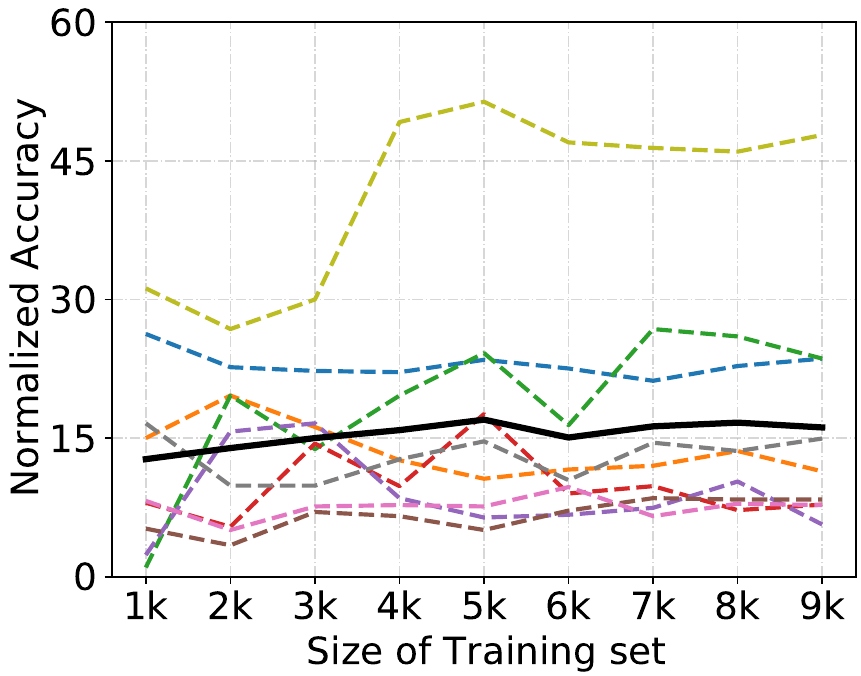}
		\label{fig:param_id1_S}
	\end{minipage}
	\begin{minipage}{0.08\linewidth}
		\centering
		\includegraphics[width=1\linewidth]{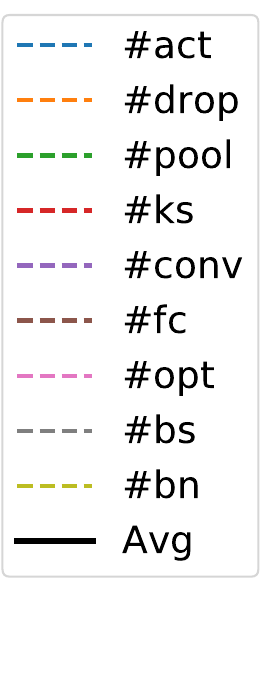}
		\label{fig:legend}
	\end{minipage}
  \vspace{-0.2in}
	\caption{Performance against size of training set on PACS modelset. From left to right, normalized accuracies in the P split, C split and S split are shown, respectively.}
	\label{fig:acc_size}
	\vspace{-0.2in}
\end{figure*}

\textcolor{black}{ \textcolor{black}{2) We conduct experiments in situations where domains are quite different, and number of classes between domains are different.}  We first train white-box models using the Sketch and Cartoon datasets. The outputs of these white-box models serve as the source domains for training DREAM. We then utilized data from CIFAR10 and CIFAR100 to train black-box models. The trained DREAM infers attributes based on the black-box model outputs. 
\textcolor{black}{The source domains (Sketch and Cartoon) comprise 7 classes, while the target domain (CIFAR10/100) contains 110 classes. We use all 7 classes as the label space for the source domains. For the target domain's label space, we select the 5 classes that overlap between the source and target domains (namely: dog, elephant, house, horse, and person).} This ensures a overlapping label space, but different number of classes between source and target domains.
The experimental results are listed in Table \ref{table:quite_diff_CIFAR}, which demonstrate that DREAM still outperforms Random, KENNEN, as well as state-of-the-art OOD generalization approaches.
This indicates that our method is effective when domains are quite different, as well as when the number of classes differs between domains.}

\begin{figure*}[t]
	\centering
	\begin{minipage}{0.31\linewidth}
		\centering
		\includegraphics[width=1\linewidth]{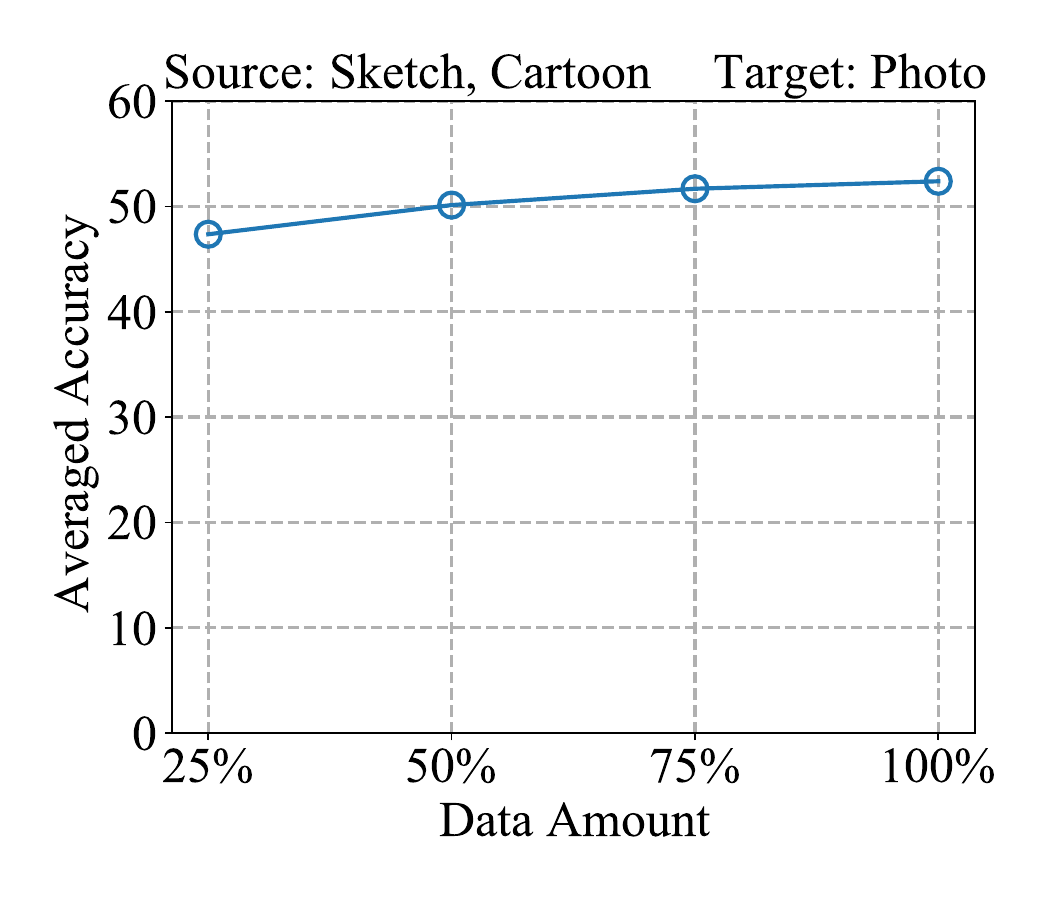}
		\label{fig:param_id2_P}
	\end{minipage}
	\begin{minipage}{0.31\linewidth}
		\centering
		\includegraphics[width=1\linewidth]{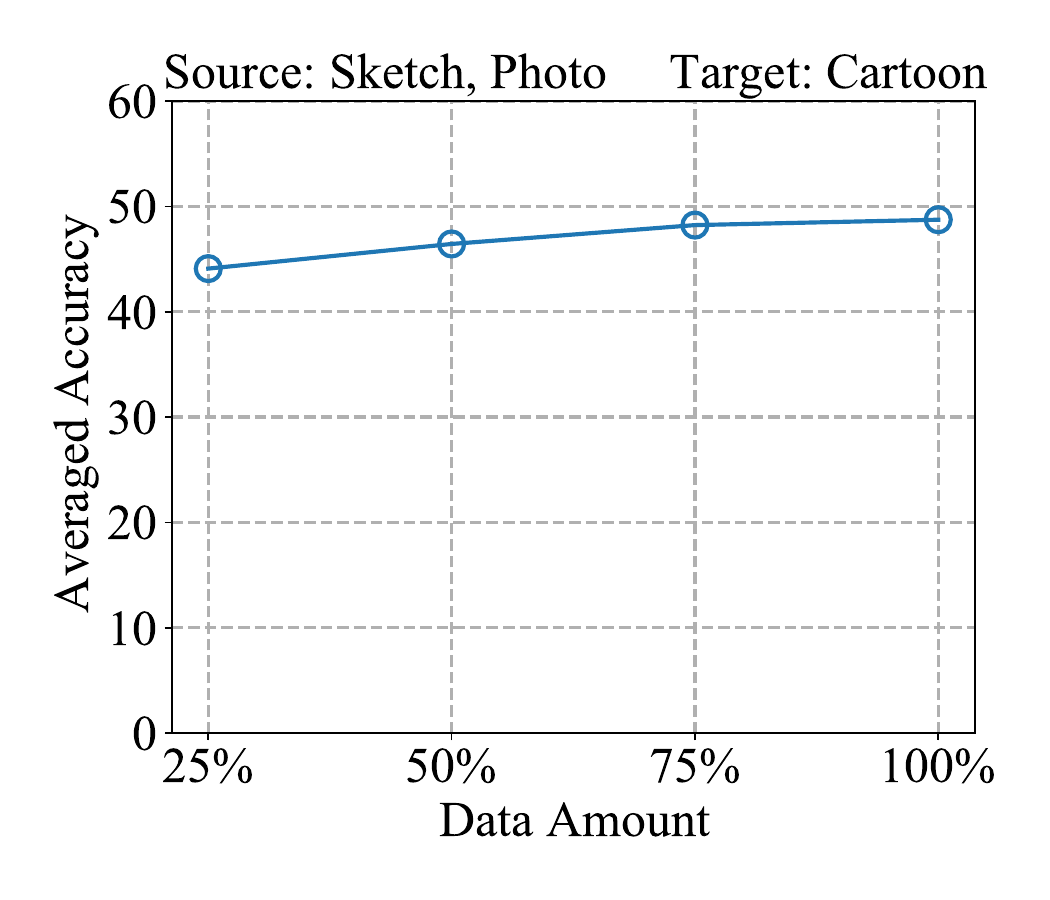}
		\label{fig:param_id2_C}
	\end{minipage}
	\begin{minipage}{0.31\linewidth}
		\centering
		\includegraphics[width=1\linewidth]{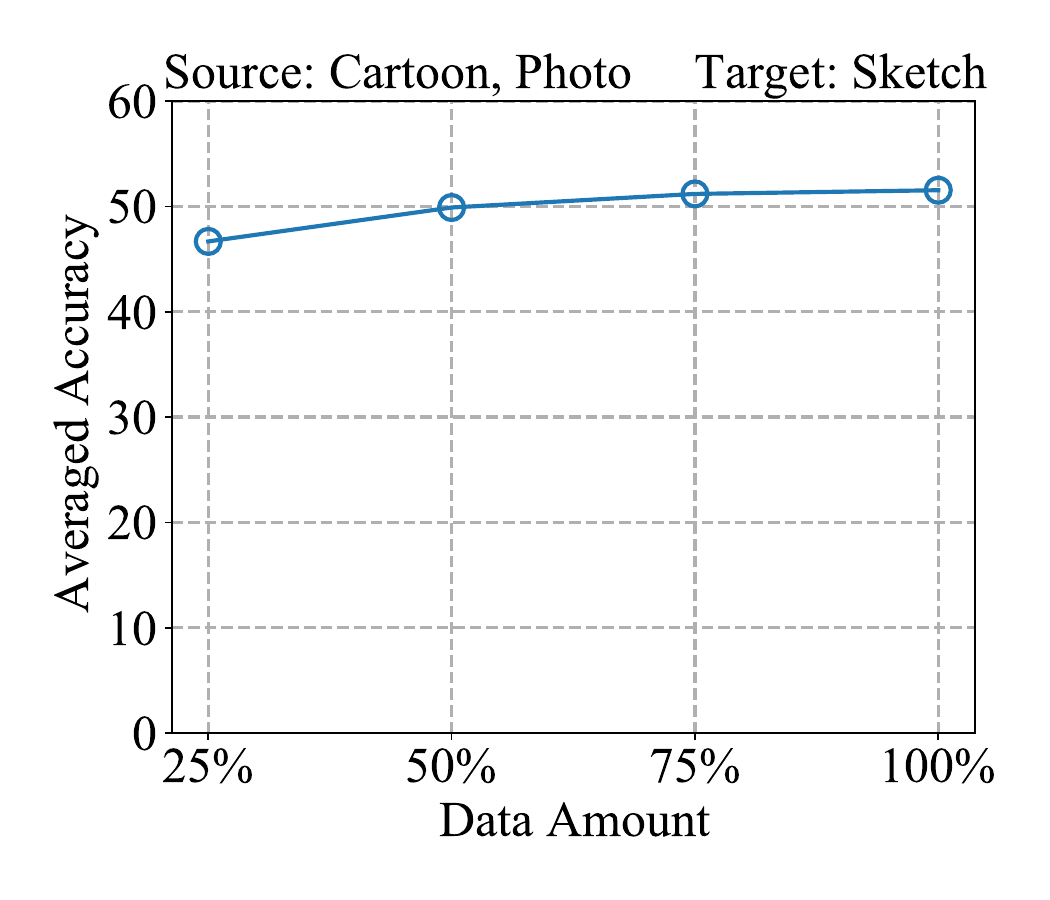}
		\label{fig:param_id2_S}
	\end{minipage}
         \begin{minipage}{0.05\linewidth}
		\centering
		\includegraphics[width=1\linewidth]{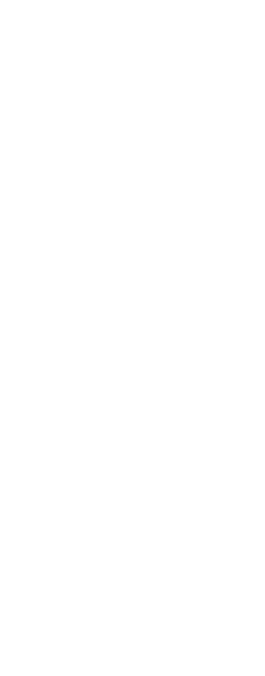}
		\label{fig:param_id2_S}
	\end{minipage}
  \vspace{-0.2in}
	\caption{Performance on data amount for reverse engineering.}
	\label{fig:data_amount}
\end{figure*}

3) We investigate the shift scenario where white-box models used for training meta-model and the black-box model to be inferred have completely different attribute combinations. As mentioned in Sect. \ref{subsec:dataset_construction}, there are a total of $5,184$ combinations of model attributes. We randomly select $3,000$, $1,000$, and $1,000$ models for the training, validation, and testing sets, respectively, ensuring that none of the models share identical attribute combinations. \textcolor{black}{We adopt the complete overlap setting.} The resulting model is denoted as DREAM**. As demonstrated in Table \ref{table:diff_arc}, DREAM** consistently outperforms other baselines under this setting. Consequently, the shift caused by attribute combinations does not significantly impact its overall performance.


\subsection{Analysis}
\textcolor{black}{We analyse the size of modelset and the training data amount. We adopt the complete overlap setting in these two experiments.}

\textbf{Analysis of Size of Modelset.}
We further study the performance of our method against the size of the PACS modelset. As shown in Fig. \ref{fig:acc_size}, we observe that the performance slightly fluctuates from the size of 1K to 5K, and does not consistently increase when the size increases. We suspect it can be attributed to the difficulty of our problem for domain-agnostic attribute inference of the black-box model, and the nature of the OOD problem, $i.e.$, the noise level increases as the size of the model set increases. It is worth studying further.

\textcolor{black}{
\textbf{Analysis of Training Data Amount.}
We analyze the impact of amount of training samples on reverse engineering performance. 
We train white-box models using 25\%, 50\%, 75\%, and 100\% of PACS training data, respectively. We then apply our proposed DREAM method to perform reverse engineering on these models. The results are presented in Fig. \ref{fig:data_amount}. We observe as the amount of data increases, the performance improves, and the performance levels off when the amount of data reaches 75\%. 
}
\textcolor{black}{
\subsection{Reverse Engineering on Model with Larger Attribute Space}
We design a larger model attribute space for model attribute inference, as shown in Table \ref{table:larger_attr}. Specifically, we
\begin{itemize}
  \item Add "GeLU" activation function to the "\#Activation" attribute space.
  \item Expand "\#Kernel size" attribute space from $\{3,5\}$ to $\{3,4,5,6,7,8,9\}$
  \item Expand "\#Conv layers" attribute space from $\{2,3,4\}$ to $\{2,3,4,5,6,7,8,9,10\}$
  \item increase "\#FC layers" attribute space from $\{2,3,4\}$ to $\{2,3,4,5,6,7,8\}$.
\end{itemize}
The outputs of models trained on the Photo and Cartoon datasets serve as the source domains in this experiment, whereas those trained on the Sketch dataset function as the target domain. \textcolor{black}{We adopt the complete overlap setting in this experiment.} The results presented in Table \ref{table:results_larger_attr} demonstrate that our method continues to outperform the baseline methods.
}
\begin{table*}[ht]
  \begin{minipage}{0.5\linewidth}
    \centering
\caption{Larger attributes space of CNN architecture.}
\vspace{-5mm}
\begin{tabular}{ll}\\
\toprule[1.3pt]
\textbf{Attribute}  & \textbf{Values} \\ \toprule [1.3pt]
\#Activation (act) & ReLU, PReLU, ELU, Tanh, GeLU \\  \midrule
\#Dropout (drop)& Yes, No  \\  \midrule
\#Max pooling (pool)& Yes, No  \\  \midrule

\#Kernel size (ks)& 3, 4, 5, 6, 7, 8, 9   \\  \midrule
\#Conv layers (conv)& 2, 3, 4, 5, 6, 7, 8, 9, 10, 11  \\  \midrule
\#FC layers (fc)& 2, 3, 4, 5, 6, 7, 8  \\  \midrule
\#Optimizer (opt)& SGD, ADAM, RMSprop   \\  \midrule
\#Batch size (bs)& 32, 64, 128  \\  \midrule
\#Batchnorm (bn)& Yes, No   \\  \midrule[1.3pt]
\end{tabular}
\label{table:larger_attr}
  \end{minipage}
  \begin{minipage}{0.5\linewidth}
    \centering
\caption{Attributes space of Vision Transformer architecture.}
\vspace{-5mm}
\begin{tabular}{ll}\\
\toprule[1.3pt]
\textbf{Attribute}  & \textbf{Values} \\ \toprule [1.3pt]
\#Activation (act) & ReLU, PReLU, ELU, Tanh, GeLU \\  \midrule
\#Dropout (drop) & Yes, No  \\  \midrule
\#Patch size (ps) & 2, 4, 8, 16  \\  \midrule
\#Transformer layers (n\_trans) & 2, 3, 4, 5, 6, 7, 8, 9,10   \\  \midrule
\#Feedforward layers (n\_ff)& 2, 3, 4  \\  \midrule
\#Attention heads (n\_heads)& 1, 2, 4, 6, 8, 12  \\  \midrule
\#Optimizer (opt)& SGD, ADAM, RMSprop   \\  \midrule
\#Batch size (bs)& 16, 32, 64  \\  \midrule[1.3pt]
\end{tabular}
\label{table:vit_attr}
  \end{minipage}
\end{table*}



\begin{table*}[t]
  \caption{Model attribute inference accuracy (\%) on outputs of Sketch models with a larger attribute space. The source domains are outputs of Photo and Cartoon models.  \textcolor{red}{\textbf{Red}} and \textcolor{black}{\underline{blue}}  indicate the best and second best performance, respectively.}
  \label{table:results_larger_attr}
  \centering
  \setlength\tabcolsep{5pt}
  \begin{tabular}{c|cccccccccccc}
  \toprule[1.3pt]
  \multirow{3}{*}{\makecell[c]{Target\\Domain}} & \multirow{2}{*}{Method} & \multicolumn{9}{c}{Attributes} & \multirow{2}{*}{Avg}  \\
  \cline{3-11}
   & & \#act & \#drop & \#pool & \#ks & \#conv & \#fc & \#opt& \#bs & \#bn \\
   \cline{2-12}
  &Random &20.00 & 50.00 & 50.00 & 14.29 & 10.00 & 14.29 & 33.33 & 33.33 & 50.00 & 30.58  \\
  \hline
  \multirow{8}{*}{Sketch}&SVM &36.32  & 58.74  & 60.91  & 17.39  & 10.80  & 17.18  & 54.53  & 35.70  & 69.24  & 40.09   \\
  &KENNEN* &35.29  & 64.61  & 68.31  & 19.75  & 15.33  & 20.99  & 62.45  & 39.40  & 69.34  & 43.94  \\
  &SelfReg &\color{blue}\underline{37.86}  & 62.65  & 66.98  & \color{blue}\underline{20.37}  & 16.46  & \color{blue}\underline{21.91}  & 66.87  & 39.09  & 66.56  & 44.31  \\
  &MixStyle &35.91  & 60.80  & 66.77  & 19.14  & \color{blue}\underline{17.49}  & 19.75  & 64.81  & \color{blue}\underline{40.43}  & 67.59  & 43.63   \\
  &MMD &36.52  & 64.61  & \color{blue}\underline{68.72}  & 19.96  & 15.64  & 19.24  & \color{blue}\underline{68.93}  & 38.89  & \color{blue}\underline{70.37}  & \color{blue}\underline{44.76}  \\
  &SD &35.08  & \color{blue}\underline{65.43}  & 65.33  & 20.06  & 15.95  & 21.50  & 63.99  & 39.40  & 67.39  & 43.79   \\
  &\textbf{DREAM} 
  &\color{red}\textbf{40.53} &\color{red}\textbf{66.77} &\color{red}\textbf{68.93} &\color{red}\textbf{25.51} &\color{red}\textbf{20.68} &\color{red}\textbf{24.79} &\color{red}\textbf{69.34} &\color{red}\textbf{44.75} &\color{red}\textbf{72.94} &\color{red}\textbf{48.25} \\
  \bottomrule[1.3pt]
\end{tabular}

\end{table*}

\begin{table*}[t]
  \caption{Model attribute inference accuracy (\%) on outputs of sketch models. The source domains are outputs of Photo and Cartoon models. The architecture is Vision Transformer. \textcolor{red}{\textbf{Red}} and \textcolor{black}{\underline{blue}}  indicate the best and second best performance, respectively.}
  \label{table:vit_results_sketch}
  \centering
  \setlength\tabcolsep{5pt}
  \begin{tabular}{c|ccccccccccc}
  \toprule[1.3pt]
  \multirow{3}{*}{\makecell[c]{Target\\Domain}} & \multirow{2}{*}{Method} & \multicolumn{8}{c}{Attributes} & \multirow{2}{*}{Avg}  \\
  \cline{3-10}
   & & \#act & \#drop & \#ps & \#n\_trans & \#n\_ffn & \#n\_head & \#opt& \#bs \\
   \cline{2-11}
  &Random &20.00 & 50.00 & 25.00 & 11.11 & 33.33 & 16.67 & 33.33 & 33.33 & 27.85  \\
  \hline
  \multirow{8}{*}{Sketch}&SVM &25.31  & 52.47  & 45.27  & 14.20  & 37.45  & 25.31  & 46.30  & 41.98  & 36.03    \\
  &KENNEN* &28.40  & 64.20  & 53.29  & 16.26  & 43.83  & 25.51  & \color{blue}\underline{51.03}  & 54.32  & \color{blue}\underline{42.10}   \\
  &SelfReg &28.19  & \color{blue}\underline{65.64}  & 40.74  & 17.08  & \color{red}\textbf{51.65}  & \color{blue}\underline{29.84}  & 42.39  & \color{blue}\underline{59.47}  & 41.87    \\
  &MixStyle &25.51  & 59.88  & 46.91  & 15.43  & 45.47  & 26.75  & 45.47  & 54.73  & 40.02    \\
  &MMD &27.57  & 56.38  & 53.29  & \color{blue}\underline{18.93}  & 48.56  & \color{blue}\underline{29.84}  & 39.09  & 56.58  & 41.44     \\
  &SD &\color{blue}\underline{29.84}  & 59.47  & \color{red}\textbf{56.38}  & 17.28  & 45.06  & 28.19  & 43.83  & 53.29  & 41.67   \\
  &\textbf{DREAM} 
  &\color{red}\textbf{34.36} &\color{red}\textbf{68.93} &\color{blue}\underline{54.73} &\color{red}\textbf{20.37} &\color{blue}\underline{49.79} &\color{red}\textbf{31.28} &\color{red}\textbf{57.00} &\color{red}\textbf{61.11} &\color{red}\textbf{47.20} \\

  \bottomrule[1.3pt]
\end{tabular}

\end{table*}

\textcolor{black}{
\subsection{Reverse Engineering on Vision Transformer Architecture}
In addition to performing attribute inference on CNN architecture, we extend our approach to Vision Transformer (ViT) architecture. The Vision Transformer (ViT) applies the Transformer architecture, originally used in natural language processing, to images. Specifically, it divides an image into several patches and uses a Transformer to extract features from the image \cite{dosovitskiy2020image}. We design an attribute space for the Vision Transformer architecture, as shown in Table \ref{table:vit_attr}. \#Patch size = $k$ means that each input image is divided into patches of size $k*k$, which are used as the input to the ViT. Additionally, we include the number of \#Transformer layers, \#Feedforward layers, and numbers of \#Attention heads in the attribute space.
In this experiment, we also use outputs from models trained on the Photo and Cartoon datasets as the source domains, and outputs from models trained on the Sketch dataset as the target domain. \textcolor{black}{The complete overlap setting is adopted.} The experimental results, listed in Table \ref{table:vit_results_sketch}, indicate that our method is capable of effectively performing attribute inference for Vision Transformer models, outperforming other baseline methods.
}


\textcolor{black}{
\section{Limitations and Future Works}
\textcolor{black}{DREAM requires training white-box models with numerous attribute combinations to ensure the performance of reverse engineering, which costs significant computational resources. For instance, when constructing the PACS modelset, each model requires approximately 5 minutes of training time. Consequently, the total training time for 13,000 models (10,000 for training, 2,000 for validation, and 1,000 for testing) amounts to roughly 45 GPU-days.} To reduce these costs, one plausible approach is to leverage fewer white-box models for reverse engineering. Considering the relationships among attributes, we can design a multi-task learning method, where reverse engineering each attribute is treated as a task, to maintain the performance in the future.
}



\section{Conclusion}
In this paper, we studied the problem of domain-agnostic reverse engineering towards the attributes of the black-box model with unknown traning dataset, and cast it as an OOD generalization problem.
We proposed a generalized framework, DREAM, which can predict the attributes of a black-box model with an unknown domain, and explored to learn domain-invariant features from probability outputs in the scenario of black-box attribute inference.
Extensive experimental results demonstrated the effectiveness of our method.

\section*{ACKNOWLEDGEMENT}
This work was supported by the NSFC under Grants 62122013, U2001211. This work was also supported by the Innovative Development Joint Fund Key Projects of Shandong NSF under Grants ZR2022LZH007.

\bibliographystyle{IEEEtran}
\bibliography{TKDE_main}
\begin{IEEEbiography}[
{\includegraphics[width=1in,height=1.25in,clip,keepaspectratio]{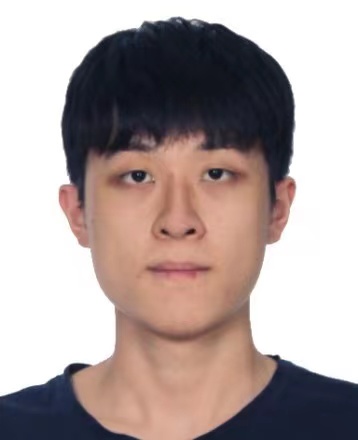}}
]{Rongqing Li} received the B.E. degree in Computer Science and Technology from Beijing University of Technology (BJUT) in 2021. He is currently pursuing the Ph.D. degree with the School of Computer Science and Technology, Beijing Institute of Technology (BIT), China. He has published several top conferences and journals, including IEEE TIP, NeurIPS, KDD and IROS. His research interests include AI security and autonomous driving.
\end{IEEEbiography}
\begin{IEEEbiography}[
{\includegraphics[width=1in,height=1.25in,clip,keepaspectratio]{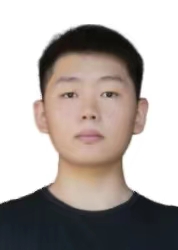}}
]{Jiaqi Yu}
received the master degree in Computer Science and Technology from Beijing Institute of Technology (BIT) in 2023. He is currently working at Kuaishou Technology as a recommendation algorithm engineer. His research interests include recommendation system, computer vision and graph neural network.
\end{IEEEbiography}
\begin{IEEEbiography}[{\includegraphics[width=1in,height=1.25in,clip,keepaspectratio]
{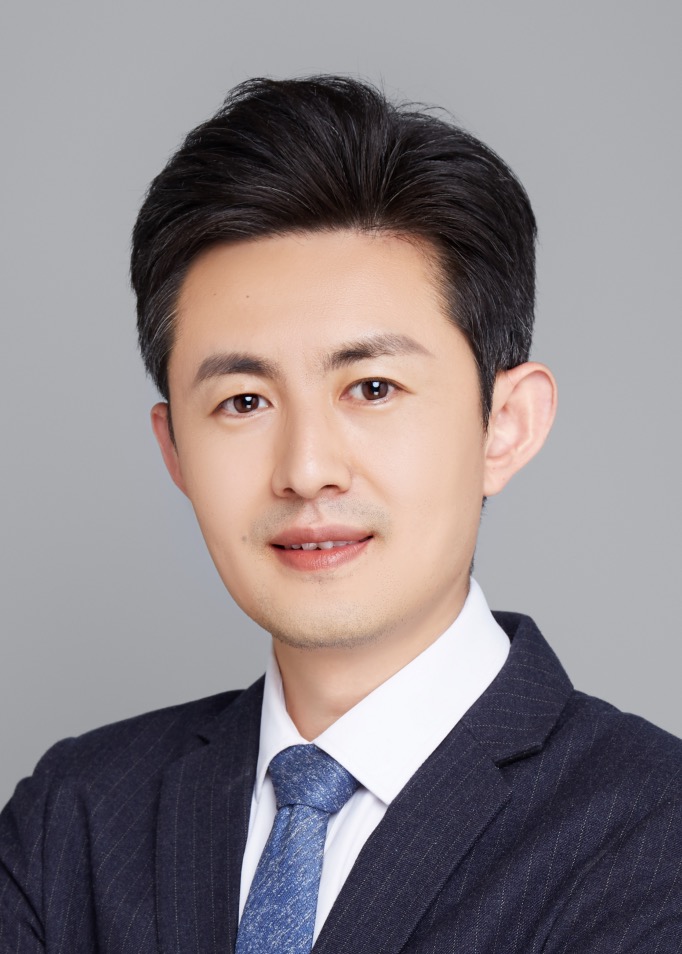}}]{Chengsheng Li}
received the B.E. degree from the University of Electronic Science and Technology of China (UESTC) in 2008 and the Ph.D. degree in pattern recognition and intelligent system from the Institute of Automation, Chinese Academy of Sciences, in 2013. During his Ph.D., he once studied as a Research Assistant with The Hong Kong Polytechnic University from 2009 to 2010. He is currently a Professor with the Beijing Institute of Technology. Before joining the Beijing Institute of Technology, he worked with IBM Research, China, Alibaba Group, and UESTC. He has more than 90 refereed publications in international journals and conferences, including IEEE TPAMI, IJCV, TIP, TKDE, NeurIPS, ICLR, ICML, PR, CVPR, AAAI, IJCAI, CIKM, MM, and ICMR. His research interests include machine learning, data mining, and computer vision. He won the National Science Fund for Excellent Young Scholars in 2021.
\end{IEEEbiography}
\begin{IEEEbiography}[
{\includegraphics[width=1in,height=1.25in,clip,keepaspectratio]{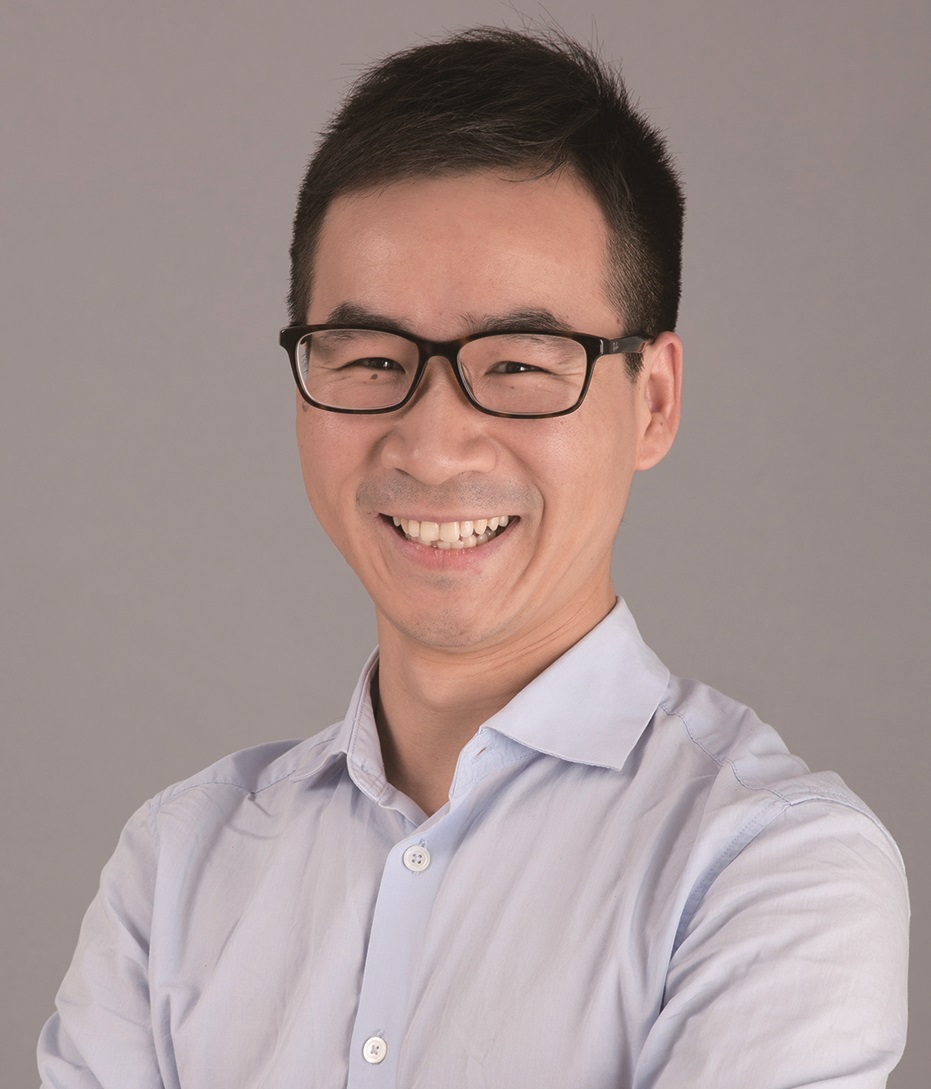}}
]{Wenhan Luo}
is currently an Associate Professor with the Hong Kong University of Science and Technology. Previously, he worked as an Associate Professor at Sun Yat-sen University. Prior to that, he worked as a research scientist for Tencent and Amazon. He has published over 90 papers in top conferences and leading journals, including ICML, NeurIPS, CVPR, ICCV, ECCV, ACL, AAAI, ICLR, TPAMI, IJCV, TIP, etc. He also has been area chair, reviewer, senior PC member and Guest Editor for several prestigious journals and conferences. His research interests include several topics in computer vision and machine learning, such as image/video synthesis, and image/video quality restoration. He received the Ph.D. degree from Imperial College London, UK, 2016, M.E. degree from Institute of Automation, Chinese Academy of Sciences, China, 2012 and B.E. degree from Huazhong University of Science and Technology, China, 2009.
\end{IEEEbiography}
\begin{IEEEbiography}[
{\includegraphics[width=1in,height=1.25in,clip,keepaspectratio]{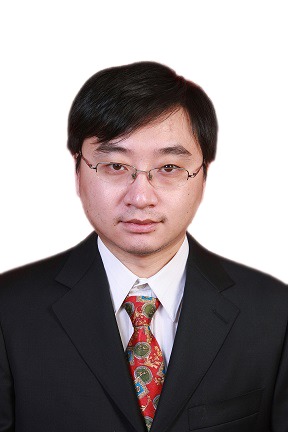}}
]{Ye Yuan} 
received the B.S., M.S., and Ph.D. degrees in computer science from Northeastern University in 2004, 2007, and 2011, respectively. He is currently a Professor with the Department of Computer Science, Beijing Institute of Technology, China. He has more than 100 refereed publications in international journals and conferences, including VLDBJ, IEEE TRANSACTIONS ON PARALLEL AND DISTRIBUTED SYSTEMS, IEEE TRANSACTIONS ON KNOWLEDGE AND DATA ENGINEERING,SIGMOD, PVLDB, ICDE, IJCAI, WWW, and KDD. His research interests include graph embedding, graph neural networks, and social network analysis. He won the National Science Fund for Excellent Young Scholars in 2016.
\end{IEEEbiography}
\begin{IEEEbiography}[
{\includegraphics[width=1in,height=1.25in,clip,keepaspectratio]{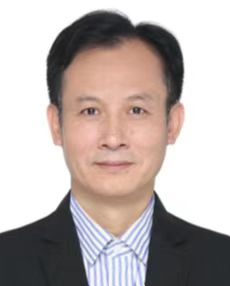}}
]{Guoren Wang}
received the B.S., M.S., and Ph.D. degrees in computer science from Northeastern University, Shenyang, in 1988, 1991, and 1996, respectively. He is currently a Professor with the School of Computer Science and Technology, Beijing Institute of Technology, Beijing, where he has been the Dean since 2020. He has more than 300 refereed publications in international journals and conferences, including VLDBJ, IEEE TRANS-ACTIONS ON PARALLEL AND DISTRIBUTED SYSTEMS, IEEE TRANSACTIONS ON KNOWLEDGE AND DATA ENGINEERING, SIGMOD, PVLDB, ICDE, SIGIR, IJCAI, WWW, and KDD. His research interests include data mining, database, machine learning, especially on highdimensional indexing, parallel database, and machine learning systems. He won the National Science Fund for Distinguished Young Scholars in 2010 and was appointed as the Changjiang Distinguished Professor in 2011.
\end{IEEEbiography}
\end{document}